\documentclass[preprint, 3p]{elsarticle}

\journal{International Journal of Forecasting}
\biboptions{longnamesfirst}
\usepackage{multirow}
\usepackage{amssymb}

\usepackage {amsmath}
\usepackage{subcaption}
\usepackage{enumerate}
\usepackage{color}
\usepackage{here}
\usepackage{comment}

\usepackage{graphicx}
\usepackage{bm}

\usepackage{upgreek}
\usepackage[normalem]{ulem}
\usepackage{tabularx}
\usepackage{stmaryrd}

\usepackage{chngcntr}
\usepackage[title]{appendix}

\usepackage{natbib}
\setcitestyle{authoryear,open={(},close={)}} 

\makeatother

\begin{document}
%
\title{PING: physics informed graph neural network for advection-diffusion processes with applications to solar power forecasting}
%


\author[1,2]{Jelena Simeunović\corref{mycorrespondingauthor}} 
\cortext[mycorrespondingauthor]{Corresponding author at:  CSEM, Rue de Jaquet-Droz 1, 2002 Neuch\^{a}tel, Switzerland.}
\ead{jelena.simeunovic@csem.ch}
\author[1] {Baptiste Schubnel}
\ead{baptiste.schubnel@csem.ch}
\author[1] {Pierre-Jean~Alet}
\ead{pierre-jean.alet@csem.ch}
\author[2] {Pascal Frossard}
\ead{pascal.frossard@epfl.ch}
\author[1] {Rafael E. Carrillo}
\ead{rafael.carrillo@csem.ch}
\address[1]{ CSEM, Rue de Jaquet-Droz 1, 2002 Neuch\^{a}tel, Switzerland}
\address[2]{ Department of Electrical Engineering, École Polytechnique Fédérale de Lausanne (EPFL), Rte Cantonale, 1015 Lausanne, Switzerland.}


\begin{abstract}

Advection-diffusion plays an essential role in transport phenomena such as atmospheric and cloud dynamics. Improved forecasting of cloud motion, in turn, enhances solar power production forecasts and facilitates the integration of additional photovoltaic (PV) sources into the power  grid. However, both classical numerical models and emerging deep learning approaches face challenges when applied to spatially irregular sensor networks such as those formed by weather stations. In this work, we adopt a graph signal processing perspective and model advection-diffusion particles as signals on a graph to capture the dynamics of these processes. We introduce a Physics-Informed Graph Neural Network (PING) for forecasting future advection-diffusion particle concentrations on irregular grids. The model incorporates a discretized advection-diffusion equation as a soft constraint in the loss function to guide learning toward physically consistent solutions. This model is on par with state of the art models on synthetic and real transport phenomena datasets,  including sea surface temperature and cloud index datasets. Furthermore, the PING model is successfully coupled with models for forecasting PV power production. This hybrid approach leverages data fusion from ground-based sensors at PV stations and satellite images with cloud concentration to improve the accuracy of intra-day PV power production.\\

\end{abstract}


\begin{keyword}
Physically informed neural network \sep machine learning \sep advection-diffusion processes \sep graph neural networks \sep photovoltaic power forecasting
\end{keyword}

\maketitle 
\section{Introduction}

Forecasting spatio-temporal processes such as cloud cover and cloud motion, ocean temperature or air pollution is a critical task in environmental and energy domains. These phenomena are governed by physical principles and could be described using advection-diffusion equations. Accurate forecast of these processes is essential in renewable energy, meteorology and other areas. For example, cloud formation and movement directly influence solar irradiance, the key factor affecting photovoltaic (PV) power production. Thus, improving cloud motion forecasts can enhance the accuracy of the PV power generation predictions, which enables better grid management and energy trading decisions. Similarly, sea surface temperature (SST) patterns affect atmospheric conditions, and improvement in SST forecasts can lead to better weather projections. However, all these processes are inherently challenging to forecast accurately using either purely data-driven or purely physical models due to their nonlinear behavior. 

Traditionally, different numerical solvers have been developed to solve advection-diffusion equations that describe these physical processes. However, these solvers are usually computationally intensive and require domain expertise to describe the process and choose physically meaningful parameters (see e.g. \cite{fortunato2022multiscale} and \cite{sanchez2020learning}). This makes purely physics-based simulations impractical for many forecasting applications, especially when rapid predictions are needed in real-time. On the other hand, data-driven models are able to learn directly from historical data and accelerate the computations. However, they are often black-box models that struggle to generalize beyond the conditions seen in the training data. Hence, when these models are faced with weather patterns that differ from training data distribution, they may yield large errors. Moreover, they tend to generate overly smooth predictions,  as discussed in \cite{de2019deep},  and fail to capture sharp changes in the dynamics of physical phenomena.

To address the above-mentioned limitations, a series of recent works have mixed approaches based on machine learning models and numerical differential equations.  The core idea is to integrate prior physics knowledge into machine learning models or to enhance traditional solvers with machine learning in order to accelerate simulations. This has led to the development of   physically informed neural networks (PINNs)  \citep{Cuomo2022ScientificML}   specifically designed to model intricate physical phenomena. In addition to accelerating simulations and enhancing the generalizability of spatio-temporal machine learning models, PINNs ensure scientific validity by enforcing adherence to physical laws  \citep{wang2021physics}. Thus, PINNs are trained not only to fit the data but also to satisfy the underlying partial differential equations (PDEs) that govern the dynamics of the processes. Most PINN studies include a term in the loss function derived from the PDE, such that the model is penalized if the network simulations deviate from the known physical equations. This encourages the neural networks to yield more plausible outputs, even when the datasets are not large. Some PINNs are used to directly approximate solutions of PDEs and they have shown success in solving both forward and inverse PDE problems, thereby eliminating the need for numerical solvers  \citep{raissi}.  Other research focuses on using PINNs to correct errors in physics-based numerical differential equations \citep{NEURIPS2021_7ca57a9f, yin2021augmenting, belbute2020combining}. Despite their rapid development in recent years, most PINN implementations are still primarily applied to the problems on a regular grid. However, many real-world phenomena come from a sensor network that inherently lies on the irregular grid, such as weather stations, air quality monitors or distributed PV stations. In these scenarios, it is not straightforward to apply PINNs, since the sensors’ locations form an irregular grid and the regular mesh topology, which is the core of standard PDE solvers, cannot be applied directly.

Graph Neural Networks (GNNs) have emerged as a natural solution for learning from data on irregular grids. The sensors could be modeled in GNN as nodes in the graph, while the edges are connecting the nodes and they encode the relationship between them. GNNs have used message-passing between the nodes, to diffuse information across nodes and to capture spatio-temporal correlations across the nodes. In recent studies they have been used to simulate physical phenomena and to forecast time series signals. The physical solvers were successfully replaced with  message-passing graph neural network models for simulations of flow dynamics in works of \cite{sanchez2020learning, pfaff2020learning}. There are also physics-informed GNN approaches that incorporate equation-based constraints  \citep{GAO2022114502},  where a graph neural Galerkin model was introduced for solving forward and inverse PDEs in a unified manner on irregular domains. These works indicate that GNN architectures informed with physical knowledge are a promising direction for forecasting the complex spatio-temporal tasks, although they face limitations in terms of complexity, including quadratic scaling with the number of nodes and over-smoothing. These issues were addressed in the works of \cite{fortunato2022multiscale, lam2022graphcast} through multi-scale approaches by using hierarchical graph approach. Furthermore, GraphCast model achieved state-of-the-art weather forecasts when it was benchmarked against numerical weather predictions (NWP) and trained on large-scale climate data  \citep{lam2022graphcast}.   Despite promising results, such as in the works of  \citep{andrychowicz2023deep, lam2022graphcast}, these methods require large datasets and computing resources, making them impractical for real-world applications. For many forecasting problems, especially in the renewable energy sector, we need to find approaches that can yield accurate predictions without large datasets, storage or computational demands.

In the solar energy forecasting field, several deep learning and GNN approaches have been used in order to improve solar power production forecasting. GNNs have been used to capture both the temporal and spatial relationships between multiple PV sites, see the works of \cite{simeunovic2021spatio, simeunovic2022interpretable} and references therein. They have used spatio-temporal attention or sequence to sequence recurrent models to capture the spatio-temporal dynamics. Similarly, in the work of \cite{YANG2025124744}, graph convolutional neural networks and spatio-temporal attention have been used to capture correlations within PV power data and temporal correlation within time-series. Another line of research, such as the work of \cite{9773879}, investigates the impact of graph connectivity on the accuracy of short-term solar forecasts. However, none of these works take into account any physical knowledge of cloud dynamics. As a result, when faced with conditions outside training distribution, these purely data-driven models yield over-smoothed forecasts and struggle to anticipate the effect of cloud movement. 

To overcome the previously mentioned challenges, we introduce a physics-informed graph neural network (PING) model\footnote{{Part of this work was presented at the 41st European Photovoltaic Solar Energy Conference (EU PVSEC), in Vienna in September 2024.}} for forecasting concentration values in transport phenomena on irregular grids  \citep{infoepfl}. 
 The proposed framework combines graph neural networks with physics-informed learning by explicitly incorporating a discretization of the advection–diffusion equation into the model and loss function. In contrast to existing approaches, PING operates directly on irregular grids and estimates historical flows in an unsupervised manner, without requiring predefined physical parameters or external flow measurements. This enables the model to capture processes such as cloud advection and diffusion in solar power forecasting and improves generalization across heterogeneous datasets and physical conditions. 
The generalization capabilities of the proposed model are demonstrated and compared across different datasets. The contributions of the current work can be summarized as follows:
\begin{itemize}
\item We introduce a physics-informed graph neural network (PING) model for forecasting particle concentration on irregular grids. A discretized advection-diffusion PDE is added as a soft constraint to the loss function to regularize the network toward physically consistent solutions. 

\item  In addition to forecasting concentrations, the proposed model estimates flow fields of the input data, which could be used for different applications, e.g. to improve the forecasting of future concentration. 

\item We evaluate the performance of the PING model across multiple domains, including cloud concentration, sea surface temperature and two different synthetic fluid-based datasets. In addition, the performance of a  PING model coupled with a previously developed PV power forecasting model is evaluated on a real PV power dataset, and shown to outperform state of the art approaches.

\end{itemize}
The rest of the paper is organized as follows: Section \ref{section_2} provides a brief overview of advection-diffusion processes and graph time series forecasting. The proposed PING architecture is explained in detail in Section \ref{section_3}. Section \ref{section_4} presents the evaluation of the PING architecture on benchmark datasets and includes its integration with the PV power forecasting model. We conclude in Section \ref{section_5}.

\section{Problem formulation}
\label{section_2}

\subsection{Time series forecasting on graphs}
We provide a brief introduction to time series forecasting on graphs. Let $\mathbf{C}^t= (C_{1}^t,... ,C_N^t)$ denote the concentration of a quantity of interest at time $t$, where the subscript $i=1,...,N$ designs points in space where this quantity is measured. We are interested in forecasting $\mathbf{C}^t$ for the next $H$ time steps given the past $M$  observations. Formally, this can be expressed as:
\begin{equation}\label{eq:Power_gsp}
\mathbf{\hat C}^t, \ldots, \mathbf{\hat{C}}^{t+H-1}=f_\beta \left( \mathbf{C}^{t-M},\dots, \mathbf{C}^{t-1}\right),
\end{equation}
where $f_\beta$ represents a parametric estimator and $\mathbf{\hat C}^t$ denotes the predicted value of $\mathbf{C}^t$ at any time t.  The parameter set $\beta$ is learned by minimizing the mean square error (MSE) across the entire forecasting horizon,
 \begin{equation}\label{eq:mse}
 \arg\min_\beta  \sum_{\tau \in \mathcal{T}}\sum_{\xi=\tau}^{\tau +H-1} \| \mathbf{\hat C}^\xi - \mathbf{C}^\xi \|^{2}_{2},
\end{equation}
where $\mathcal{T}$ denotes the historical time step used for the training.

\subsection{Graph definitions}

We represent the problem using a graph $G=\{ \nu, \upvarepsilon, \mathbf{A} \}$, where $\nu$ is the set of nodes or vertices where $C_{i}^t$ is measured, $\upvarepsilon$ is the set of the edges that define if two nodes are connected and $\mathbf{A}$ denotes the adjacency matrix that describes the topology of the graph. Each adjacency entry $A_{ij}$ represents an edge between nodes $\nu_i$ and $\nu_j$. If there is no edge between $\nu_i$ and $\nu_j$ in $G$, then the matrix entry $A_{ij}$ is zero. The adjacency matrix can be unweighted or weighted, where the weights define a relation between nodes, e.g. a function of the distance between two nodes. In the unweighted adjacency matrix, the adjacency entry $A_{ij}$  is equal to 1 if there is a connection between nodes $\nu_i$ and $\nu_j$. The symmetric normalized Laplacian matrix $\mathcal{{L}}$ is defined as: $\mathcal{{L}}:= I - \mathbf{D}^{-1/2}\mathbf{A}\mathbf{D}^{-1/2}$, where $\mathbf{D}$ is the diagonal matrix of nodes' degrees $ D_{ii}=\sum_j A_{ij}$ and $I$ is the identity matrix of the appropriate dimension. We use this unweighted Laplacian later to restrict the information passing to the neighbours. However, in order to encode the underlying physical distance between nodes, we use a weighted adjacency $\mathbf W$ with entries $w_{ij}$, using the RBF kernel to define edge weights:
\begin{equation}
w_{ij} = \exp(- \frac{||\nu_i - \nu_j||^2}{2\sigma^2}),
\end{equation}
where $||\nu_i - \nu_j||$ is the Euclidean distance between the two nodes and    $\sigma$ is a kernel bandwidth (or length scale). This parameter governs how quickly  $w_{ij}$ decays as a function of the Euclidean distance between the nodes $\nu_i$ and $\nu_j$, since it depends on the mean of Euclidean distances $\bar d$ between each node and its neighbours. The multiplicative constant of the bandwidth is typically chosen experimentally. 
The weighted adjacency encodes the distance-based locality and has proven to be successful in previous works for predicting transport phenomena for sea surface temperatures \citep{de2019deep}. The corresponding weighted degree matrix and weighted graph Laplacian are obtained simply in the exact same way, as previously defined for the degree matrix and Laplacian, by simply replacing the unweighted adjacency $\mathbf{A}$ for weighted matrix $\mathbf{W}$. For a comprehensive review of graph signal processing, we refer the readers to the literature, e.g. to the works of \cite{8347162}.

Multiplying the graph signal $\mathbf C$ with the Laplacian $\mathbf {\mathcal L}$ yields: 
\begin{equation}\label{eq:laplacian_x_signal}
(\mathcal L \mathbf C)(\nu_i) = \sum_{j \in \mathcal N_i} \mathcal L_{ij} C_j = \sum_{j \in \mathcal N_i} w_{ij} (C_i - C_j),
\end{equation}
where $w_{ij}$ is the weighted matrix entry and $\mathcal N_i$ the neighbours of node $\nu_i$. Furthermore, we can define the Laplacian quadratic form that measures how much the signal changes between nodes:
\begin{equation}\label{eq:dirichlet}
\mathbf{C}^T \mathcal L \mathbf{C} = \sum_{i=1}^N \sum_{j=1}^N w_{ij} \bigl(C_i - C_j\bigr)^{2}.
\end{equation}
The graph Laplacian encodes the connectivity and structure of the graph. As such, its properties govern the discrete diffusion processes on graphs. The Laplacian is a positive definite matrix and a self-adjoint operator, thus, its factorizations $\mathcal L= \mathbf{B}^T\mathbf{B}$ can be used to define the gradient and divergence operators of functions defined on a graph. To make the relationship between the two, we can use the identity $\nabla^2 =\nabla \cdot \nabla$ in the continuous domain and note that the divergence is the adjoint of the gradient. In the graph domain, the matrix $\mathbf{B}$, also called the incidence matrix, is analogous to the gradient that operates from the vertex (node) domain to the edge domain and takes the difference between the graph signal at two nodes. Its transpose matrix, $\mathbf{B}^T$, is the adjoint operator, i.e. the analogous to the divergence in the graph domain, that operates from the edge domain back to the vertex domain by aggregating the contribution of all edges incident to a node. Thus, in the graph domain we can define the gradient of $\mathbf{C}$ along the edge $\upvarepsilon_{ij}$ between nodes ${\nu_i}$ and ${\nu_j}$ as
\begin{equation}\label{eq:difussion}
(\nabla C)(\upvarepsilon_{ij}) = \mathbf{B}\mathbf{C}= \sqrt w_{ij} (C_i-C_j),
\end{equation}
and the divergence of the gradient at node ${\nu_i}$ as the weighted sum over all edges incident to ${\nu_i}$, thus
\begin{equation}
(\nabla \cdot\nabla C)(\nu_i) = \mathbf{B}^T\mathbf{B}\mathbf{C} = \sum_{j \in \mathcal N_i} w_{ij} (C_i - C_j).
\end{equation}

 \subsection{Discretization of the advection-diffusion equation on an irregular grid}
\label{section_2c} 
Advection-diffusion describes the transport of a scalar property in a medium that follows the differential equation:
 \begin{equation}\label{eq:advdiff1}
\frac{\partial C}{\partial t}  + {\mathbf  \nabla} \cdot (\mathbf u C) =  {\mathbf  \nabla}\cdot (D {\mathbf  \nabla} C) + P,
\end{equation}
where $C$ is  the scalar property of interest, e.g. particles mass concentration, $P$ sinks and sources,  $\mathbf u$ the velocity field,  $D$ the diffusion coefficient and ${\nabla} \cdot $ the divergence operator.  For ease of notation and simplicity, we will refer to the quantity of interest (the mass concentration) simply as the concentration.  The advection describes the bulk motion of the fluid, transporting particles with concentration $C(\mathbf{x},t)$ with the velocity of fluid $\mathbf u(\mathbf {x}, t)$, over a period of time. In contrast, the diffusion occurs as particles move from area of higher concentration to lower concentration. For further details on advection-diffusion, see e.g. \cite{stocker2011introduction}.  

We discretize this equation in space and time for concentrations sitting on graphs with discrete time steps. Furthermore, we define the temporal change in concentration at node $\nu_i$ as $C_i^t-C_i^{t-1}$. We can discretize the diffusion term by replacing the continuous Laplacian operator $\nabla^2 C$ with the graph Laplacian $\mathcal L \mathbf C$ defined in equation \eqref{eq:difussion}. 


Discretizing the advection term requires additional effort though.  In this work we model different quantities as purely 2D advection-diffusion processes that lie in the horizontal plane. This assumption holds for the scenarios when vertical motions are weak or when the quantity of interest varies mostly on the horizontal scales.  Thus, the velocities $\mathbf u_i^t$ in Equation \eqref{eq:advdiff1} for the 2D advection-diffusion equation are two-dimensional. In order to represent them on a graph, we define the projection of the velocity vector $\mathbf  u_i^t$ onto the edge direction between a node $\nu_i$ and its neighbor $\nu_j$ as $ \mathbf u_i^t \cdot \mathbf e_{ij}$, where $ \mathbf {\mathbf e_{ij}} =  \frac{\Delta \mathbf z_{ij}}{|| \Delta \mathbf z_{ij} \Vert_2 ^2}$ and  $\mathbf{\Delta z}_{ij}$ is the spatial distance $ \mathbf{\Delta z}_{ij} = [x_i - x_j, y_i - y_j]$ between the nodes. We can then rewrite the advection term at node $\nu_i$ and time $t$  in a graph perspective as:
\begin{equation}\label{eq:irr_velocity0}
\nabla  \cdot \left( \mathbf{u}  C \right)_i^t := \sum_{j \in \mathcal N_i}   \sqrt {w_{ij}}  \mathbf u_i^t \cdot  \mathbf e_{ij} (C_j^t - C_i^t).
\end{equation}
Putting all terms together and assuming no sources and sinks ($P=0$ in Equation \eqref{eq:advdiff1}), we replace Equation \eqref{eq:advdiff1} on a graph by 
\begin{equation}\label{eq:irr_velocity}
C_i^{t+1}-C_i^{t} = \sum_{j \in \mathcal N_i} \sqrt {w_{ij}}\mathbf u_i^t \cdot  \mathbf e_{ij}   (C_i^t - C_j^t) +D\mathcal L_i  \bf C^t,
\end{equation}
for all $i =1, ..., N$.  
We assume in this work that the fluid is incompressible and that the velocity field satisfies $\nabla \cdot u = 0$. Thus, we can simplify the term $\nabla \cdot (\mathbf u C )$ to $\mathbf u \cdot \nabla C$. This allows us to discretize the divergence of the velocity field by projecting the velocity vector from two-dimensional space onto the direction of edge between the observed node and its neighbour. Thus, this leads us to approximate the divergence of the velocity field on an irregular grid between node $\nu_i$ and its neighbours as
 \begin{equation}\label{eq:PDE_v}
\nabla \cdot  {\mathbf u}_{i}^t \approx \sum_{j \in \mathcal N_i} \sqrt {w_{ij}} \left( (\mathbf u_i^t  - \mathbf u_j^t) \cdot  \mathbf { e_{ij}} \right), 
\end{equation}
for all $i =1, ..., N$.

\section{Physically-informed graph neural network}
\label{section_3}
\begin{figure}[h!]
  \centering
   \includegraphics[trim=0cm 7cm 0cm 1cm, clip, width=\linewidth]{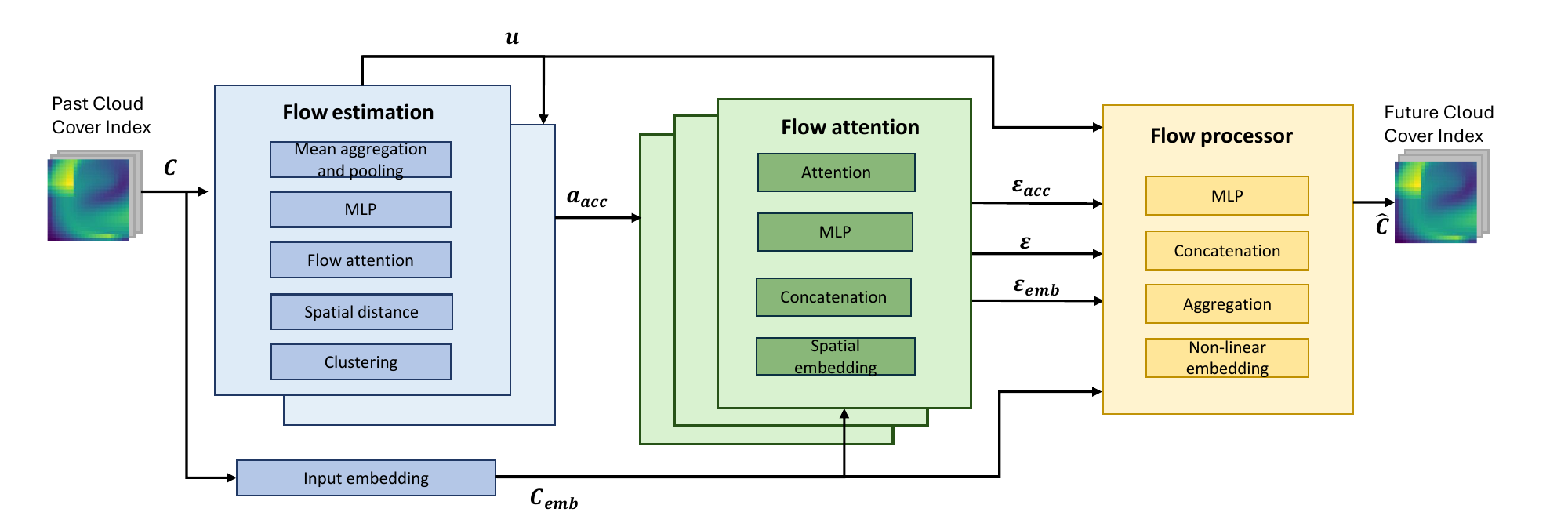}
    \caption{ PING model with three main parts: flow estimation, flow attention (edge estimation) and flow processor (flow aggregation). The flow estimation part has as input past concentration values $\bf{C}$ and outputs past velocity and acceleration field estimates. The flow attention takes as inputs the estimated past velocity field $\bf{v}$ and acceleration fields $\bf{a_{cc}}$ as well as the embedded input concentrations $\bf{C_{emb}}$. The Flow processor predicts the future concentration values $\bf{\hat C}$ using the edge values from the second block. When used for PV power prediction, the predicted concentration (cloud concentration in that case) and future clear sky values $\bf{Y}$ are fed to the decoder of the PV power forecasting model.}
    \label{fig:PING_overview}
\end{figure}

The physically-informed graph neural network (PING) proposed here is a sequence-to-sequence physics-guided model. PING is designed to predict 2D-particle concentration  phenomena that are governed by advection-diffusion processes.  An overview of the model is illustrated in Figure \ref{fig:PING_overview}. PING has three main blocks: a flow estimator, a flow attention mechanism and a flow processor block.  
The flow estimation block estimates the velocity $\mathbf u$ and acceleration values $\mathbf a_{cc}$ at each node. The flow attention block builds a graph and computes the edge direction vectors between two nodes, on which we later project velocity values. This block actually computes the  physics-based attention by querying the velocities, using the edge direction weights as keys and finally the neighboring concentrations are used as values. The processor block aggregates the information coming from different flow attention blocks and predicts the future concentration.
The estimated velocities, concentrations, and diffusion constant from these blocks are incorporated into additional loss terms in the loss function that act as soft constraints to guide the model. 
The model takes as input the sequence of the last past $M$ measurements of particle concentration $\mathbf C \in \mathbb{R}^{N \times M}$, across  $N$ nodes, and forecasts the concentration for the next $H$ steps, $\mathbf{\hat {\mathbf C}} \in \mathbb{R}^{N \times H}$, over the same set of nodes.

Each block is described in details in the next sections.

\subsection{Flow Estimator}
\label{flow_estim}
\begin{figure}[h!]
  \centering
   \includegraphics[trim=7cm 1cm 6cm 3cm, clip, width=0.7\linewidth]{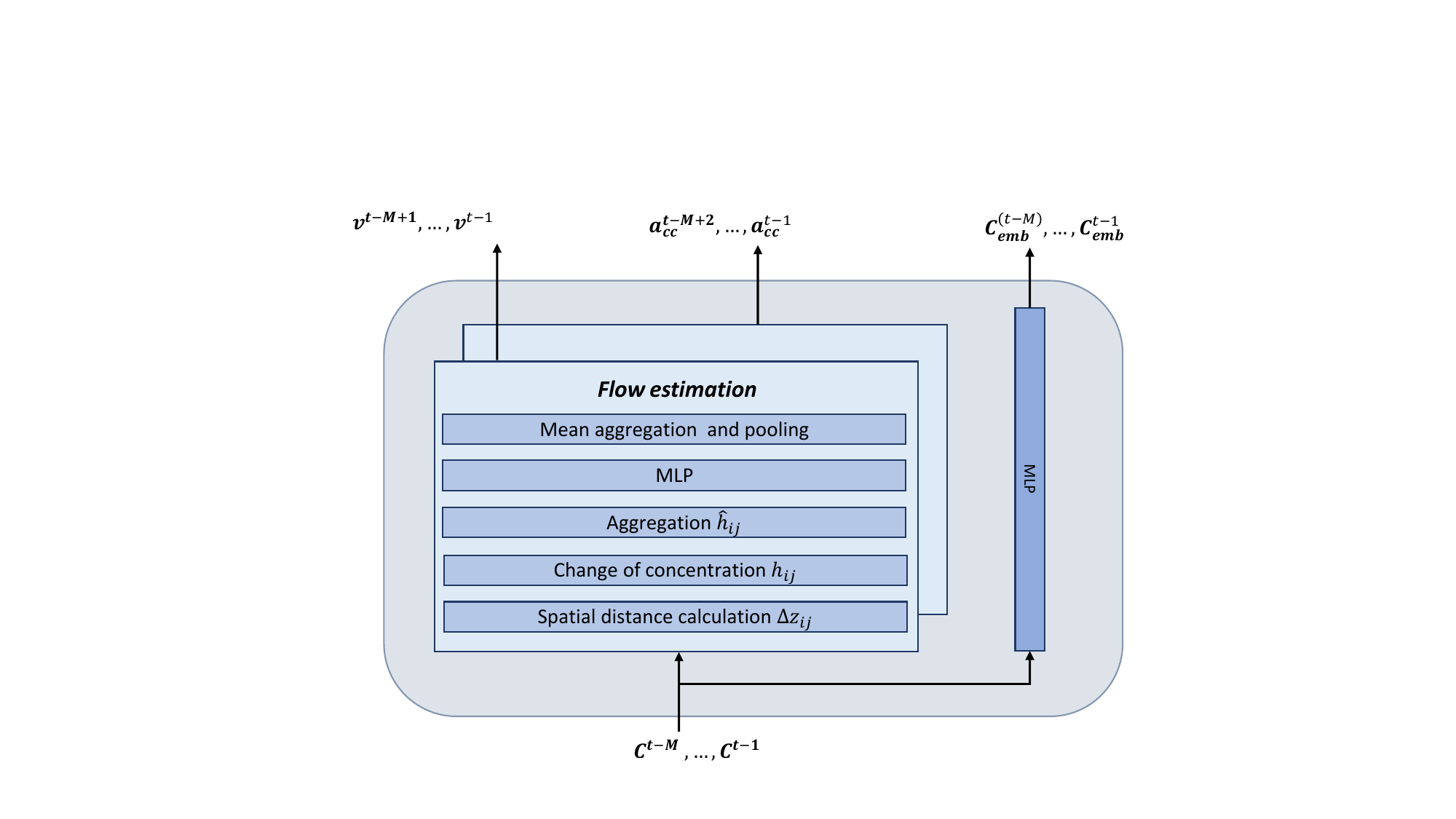}
    \caption{The flow estimator block of the PING model. }
    \label{fig:PING_encoder}
\end{figure}

The flow estimator shown in  Figure \ref{fig:PING_encoder} estimates the past velocity and acceleration fields using historical concentration values. These velocity and acceleration features are subsequently fed into the flow attention block; see Figure \ref{fig:PING_overview}. Inspired by the work of \cite{boussif2022magnet}, our model reconstructs the velocity field using Equation \eqref{eq:irr_velocity}. This equation represents the generalization of the 2D optical flow equation with an additional diffusion term.  The first layer (see Figure \ref{fig:PING_encoder}) computes the spatial difference $ \mathbf{\Delta z}_{ij} = [x_i - x_j, y_i - y_j] $ between two nodes  ($ \mathbf{\Delta z}_{ij}$ is defined in  Section \ref{section_2c}).  The ``change of concentration`` layer depicted on Figure \ref{fig:PING_encoder} computes the spatial concentration differences $C_i^t - C_j^t$,    (first term on the right side of Eq. \eqref{eq:irr_velocity}). This is further concatenated with the change of concentration over time $C_i^t -  C_{i}^{t-1}$ (left-hand side of Eq. \eqref{eq:irr_velocity}) and the diffusion term (second term on the right-hand side of Eq. \eqref{eq:irr_velocity}). They  form  together the feature vector 
 \begin{equation}\label{eq:1}
\mathbf h_{ij}^t := \left[ {\mathbf { \Delta z}_{ij}} || ( C_i^t -  C_{i}^{t-1}) || ( C_i^t -  C_{j}^{t}) || D\mathbf {\mathcal L}_i \mathbf{C}^t  \right], \quad \mathbf h_{ij}^t \in  \mathbb{R}^{F_{in}},
\end{equation}
  where the concatenation is denoted with $||$.
Using Eq. \eqref{eq:irr_velocity}, $\mathbf h_{ij}^t$ can be used to estimate the velocity field   ${\mathbf u}_{i}^t$.
 
The aggregated information across all neighbors of a given node is computed in the ``Aggregation`` layer. A learnable weighted coefficient matrix, $\mathbf W \in \mathbb{R}^{N \times S \times \frac{t}{2}}$, is used to generate a node-specific aggregated feature from neighbors,
 \begin{equation}\label{eq:2}
\mathbf {\hat h}_{i}^t= \sigma \left( \sum_{j \in \mathcal N_i} W_{ij}^t f(\bm h_{ij}^t) \right),
\end{equation}
where the total number of nodes, $S = |\mathcal{N}_i|, \forall \nu_i$ is a fixed number of the closest neighbors defined for all nodes, $ f(\cdot)$ is an MLP and $\sigma$ is an activation function.  These features are then processed by another MLP layer, $g(\cdot)$, to produce velocity features $ \hat {\mathbf u}_i^t \in \mathbb{R}^{2}$ at each node $\nu_i$, mapping them back to 2D space. To diffuse information across time, a 1D temporal convolution is applied, followed by a rolling mean aggregation over the feature sequences, resulting in the final velocity-like vector ${\mathbf {v}}_i^t \in \mathbb{R}^{2}$.

The estimated velocity features ${\mathbf v}_i^{t} $ are fed into a subsequent block to derive acceleration features ${\mathbf a}_{acc}^{t} $, as illustrated in Figure \ref{fig:PING_encoder}. 
   The same operations used to estimate velocity are applied in the feature space to compute high-dimensional acceleration features. These are referred to as acceleration features because they are derived from processing the velocity estimates, although they do not correspond to physical acceleration vectors. While the velocity and acceleration features capture the influence of neighboring nodes for flow prediction at node $\nu_i$, it is also essential to propagate the concentration input signal. To this end, the concentration input is embedded into $\mathbf C_{emb}^t$ using a single MLP layer, and is then passed—together with the velocity and acceleration features—into the Flow Attention block.

\subsection{Flow Attention}

The flow attention module consists of three attention blocks. These blocks are designed to compute the edge values that are subsequently used in the Flow aggregation layer; see Figure \ref{fig:PING_overview}.  Each of these blocks processes one of the following specific inputs: the embedded concentration values $\mathbf{C}_{emb}^t$, the estimated velocity features $\mathbf{v}^t = \left[ \mathbf{v}_1^t, \ldots, \mathbf{v}_N^t \right]$ (where $N$ denotes the number of nodes), and the acceleration features $\mathbf{a}_{acc}^t$.

 We only describe in detail what happens in the velocity attention head for the sake of simplicity; similar operations are carried out in the two other heads. The positions of the nodes are first embedded into a high-dimensional feature space through an MLP. They are represented as $ \tilde{\mathbf z_i}=\sigma(\mathbf z_i \mathbf W^{emb})$. Next, the difference between the embeddings of each pair of neighboring nodes is computed. The obtained feature vectors $\Delta \tilde{\mathbf z_i}$  are then concatenated with the velocity field estimates of the neighboring nodes  to form the feature vector 
\begin{equation}\label{eq:17}
\bm g_{ij}^t = \left[   {\mathbf v}_{j}^t  \vert  \Delta \tilde{\mathbf z}_{ij}  \right], \quad \mathbf g_{ij}^t \in \mathbb R^{F_{out} + d^{emb}},
\end{equation}
for all nodes $i$ and its neighboring nodes $j$.  The velocity vectors, which carry the dynamical information, are concatenated with positional embeddings, which provide positional context. This concatenation is performed to propagate signals on irregular grids in a manner that mimics the underlying physics, as in the works of \cite{fortunato2022multiscale}. These combined features are then passed through an MLP and projected into a one-dimensional space to compute the attention coefficients that yield the edge attention values.  The attention map uses the softmax function and is defined by
\begin{equation}
\label{eq:edges}
\varepsilon_{ij}^t = \frac{\exp \left( \beta p_{ij}^t\right)}{\sum_{k \in \mathcal N_i} \exp \left( \beta p_{ik} \right) }, 
\end{equation}
where $ { p_{ij}^t}$ is the one-dimensional projection obtained from the feature vector $\bm g_{ij}^t$ after the MLP. The inverse temperature coefficient $\beta$ is a learnable parameter. Similar operations are carried out in the two other heads for the concentration and acceleration features, replacing the velocity vectors in Equation \eqref{eq:17} by $\mathbf C_{emb}^t$ and $\mathbf a_{acc}^t$, leading to the edge features represented in Figure \ref{fig:PING_overview}.

\subsection{Flow Processor}

The flow processor block processes  estimated feature vectors and edge values to generate future concentration values. Its inputs include the embedded concentration values $\mathbf C_{emb}^t$, the velocity features $\mathbf v^t $, the  acceleration features $\mathbf a_{acc}^t$, and the attention weights corresponding to each of them
Here again, we only describe in detail the operations for the velocity features, similar operations being applied to the acceleration and concentration features. For every node $i$, the velocity features of its neighbors are aggregated using the edge features from the flow attention block 
\begin{equation}
\label{res1}
{\hat{\mathbf y}_i}= \sum_{t} \sum_{j \in \mathcal{N}_i} \sigma (\mathbf  W_{agg} {\mathbf v}_{j}^t ) \varepsilon_{ij}^{t},
\end{equation}
where $ {\hat{\mathbf y}_i}\in \mathbb R^{ Hq} $,  $\sigma$ is a leaky-ReLU activation function and $\varepsilon_{ij}^{t}$ are edge weights computed in the attention block in Eq. \ref{eq:edges}.
The final predictions of the concentration are obtained with an MLP $f^{fin}$, whose inputs are the embedded concentrations, velocity and acceleration features, denoted with ${\hat{\mathbf y}}_i^{ in}, {\hat {\mathbf y}}_i^{vel}, {\hat {\mathbf y}}_i^{acc}$, respectively:
\begin{equation} \label{final_C}
\hat{ C}_i^{t}, ...,\hat{ C}_i^{t+H-1} = f^{fin} \left[{\hat {\mathbf y}}_i^{in} || {\hat {\mathbf y}}_i^{vel} || {\hat {\mathbf y}}_i^{acc}\right].
\end{equation}

\subsection{Physics-guided loss function}
To ensure that the proposed model learns concentration and velocity estimates consistent with physical laws, two additional terms are incorporated into the MSE loss defined in Eq. \eqref{eq:mse}.  The first term of the equation forces the model to minimize the error between the forecasted and observed concentration. The second term ensures that the network adheres to the advection–diffusion equation by penalizing deviations from it and guiding the model towards physically plausible solutions. The third term enforces divergence constraints on the velocity field, as defined in Eq.~\eqref{eq:PDE_v}.   Consequently, the loss function used for training the model is then given, in continuous domain, by
\begin{equation}\label{eq:pde_2reg}
\begin{split}
L(\theta) &= \sum_{\tau \in \mathcal{T}}\sum_{\xi=\tau}^{\tau +H-1} \Big\| \mathbf{\hat C}(\xi) - \mathbf{C}(\xi) \Big\|^{2}_{2} \\
&\quad + \lambda_1 \sum_{\tau \in \mathcal{T}}\sum_{\xi=\tau-M+1}^{\tau -1} \Big\| - \frac{\partial {\mathbf C}(\xi)}{\partial t} - {\nabla} \cdot ( \mathbf u(\xi) \mathbf C(\xi)) \\
&\quad + {\nabla} \cdot (D {\nabla} \mathbf C(\xi)) \Big\|^{2}_{2} 
+ \lambda_2 \sum_{\tau \in \mathcal{T}}\sum_{\xi=\tau-M+2}^{\tau -1} \Big\| {\nabla} \cdot \mathbf u(\xi) \Big\|^{2}_{2},
\end{split}
\end{equation}
where $\mathcal{T}$ denotes the times of historical time steps utilized for fitting the model during training and D is a  learnable  hyperparameter representing the diffusion constant. In the objective function $L(\theta)$, the concentration values $C(\theta)$ and the velocities $\mathbf{u}(\theta)$ are functions of the learnable parameters $\theta$, which represent all the model parameters (learnable weights, biases, ...) described in previous sections.    The parameters $\lambda_1$ and $\lambda_2$ were determined empirically and kept relatively small, as larger values tend to oversmooth the velocity field and degrade the accuracy of the concentration forecasts. Although the three components are not on the same scale, empirical tuning of $\lambda_1$ and $\lambda_2$ allowed us to balance physical consistency with predictive accuracy.   In practice, the models are trained using stochastic gradient descent, with the sum in \eqref{eq:pde_2reg} evaluated over mini-batches, and the continuous differential equations approximated by their discrete counterparts defined in Equations \eqref{eq:irr_velocity} and \eqref{eq:PDE_v}, using the RBF kernel for the edge weights.

\section{Results}
\label{section_4}
The PING model described in Section \ref{section_3} was compared against baseline models on two types of datasets: synthetic datasets generated through flow simulations  where the model’s unsupervised velocity estimation could be precisely analyzed, and real-world datasets, which include sea surface temperature and cloud concentration indexes. At the end of this section,  the performance of the PING model integrated with a GCLSTM model was tested on a real PV power forecasting dataset used in \cite{simeunovic2021spatio}.

\subsection{Datasets}
Two simulated synthetic datasets were generated using the PhiFlow\footnote{{https://github.com/tum-pbs/PhiFlow}} package, \citep{holl2020learning}. The simulations were based on advection-diffusion equations while ensuring fluid incompressibility.  The PhiFlow model was used to generate two synthetic datasets, since the simulated smoke flow in synthetic datasets could be used as a proxy for cloud movement due to underlying similar advection-diffusion dynamics. One dataset was simulated with purely advective behavior ($D=0$), while the other incorporated diffusion and it was simulated with the diffusion constant from the PhiFlow package of $D=0.6$. We selected simulations with sources that emitted smoke at the border of the simulated region. This choice was motivated by the analogy to the real application of cloud motion forecasting: when monitoring clouds from ground stations or satellites, new clouds usually enter the observed domain from the boundaries rather than spontaneously forming in the center of the field of view. In the synthetic datasets, the smoke is introduced at 4 sources. The inflows are perturbed with small Gaussian noise ($\sigma$ = 0.005) and change every 50 steps, while the simulation evolves over 30,000 steps with states saved every 1,000 steps. The smoke is transported using a semi-Lagrangian advection scheme with buoyancy and incompressibility constraints.  For both datasets, $24 000$ time steps were simulated for training and $6000$ for evaluation. The time step $\Delta t$ was set such that the smoke density moved one spatial unit per time unit. Every $50$ steps, smoke was injected at random locations. These datasets were defined on a regular grid with a window size of $22 \times 22$. To test the models on irregular grids, four configurations of $150$ nodes ($S1, S2, S3,$ and $S4$) were randomly sampled from the regular grid.

The real-world datasets used in this study included records of sea surface temperature and cloud coverage, both sourced from ERA5\footnote{{https://cds.climate.copernicus.eu/}}, a global atmospheric reanalysis dataset produced by the European Centre for Medium Range Weather Forecasts (ECMWF). The cloud coverage dataset had an hourly resolution, while the sea surface temperature dataset  consisted of daily measurements. Both datasets have a spatial resolution of 0.25° latitude-longitude, and experiments were conducted on a $20\times 20$ window on the regular domain. The training period for cloud data spanned from 2013 to 2016, while the sea surface temperature dataset ranged from 2006 to 2017. Evaluation was conducted on the year 2017 for the cloud dataset and from 2018 to 2021 for the sea surface temperature  dataset. For regular grid evaluations, 400 nodes were used, while irregular grid tests were conducted using the same 150-node configurations sampled from the datasets.
For all datasets, the same nodes were used as inputs and forecasting nodes.
 Another real-world dataset used in this study is a PV power forecasting dataset of 304 PV stations spread inhomogeneously across Switzerland. This dataset has 15-minutes resolution and the data is split into training data over the full year of 2016 and evaluated on the year 2017. Further details about the dataset can be found in \cite{simeunovic2021spatio}. This dataset was selected because it represents real PV power generation under variable and cloudy conditions. In contrast, synthetic PV datasets exhibit smoother signals, where the impact of cloud dynamics and the forecasting errors due to cloud dynamics are less pronounced compared to real-world scenarios. 

We chose four different types of subsampling patterns for all four datasets that reside on the regular grid: cloud cover index, SST, advection and advection-diffusion datasets. For each of these types of data, the points are  subsampled in the same manner. They are subsampled randomly three times and once regularly, creating four different sets of locations with a  size of 150 points, shown in Figure \ref{fig:graph345} below. 
\begin{figure}[h!]
  \centering
  \begin{subfigure}[b]{0.4\linewidth}
    \includegraphics[trim=0.5cm 0cm 4cm 2cm,clip,width=\linewidth]{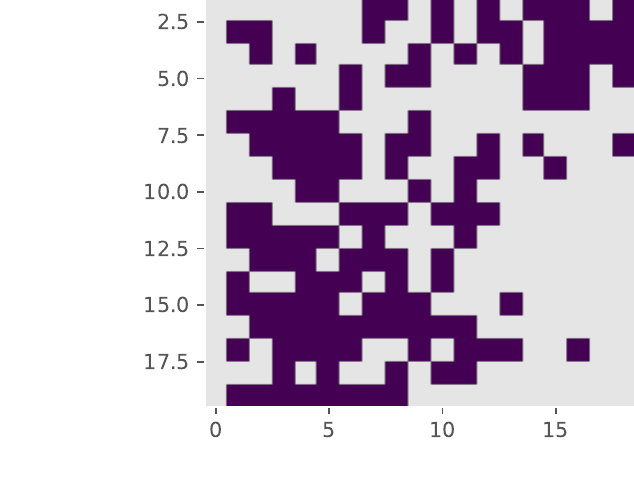}
    \caption{}
    \label{fig:graph1}
  \end{subfigure}\quad
  \begin{subfigure}[b]{0.4\linewidth}
    \includegraphics[trim=0.5cm 0cm 4cm 2cm,clip,width=\linewidth]{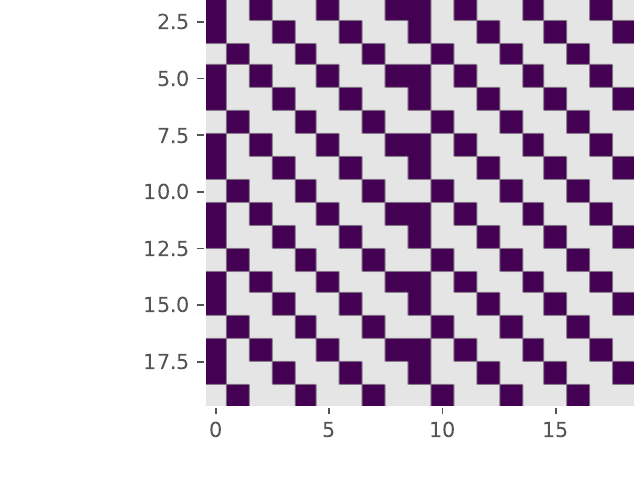}
    \caption{}
    \label{fig:graph2}
  \end{subfigure}
  
  \medskip 
  
  \begin{subfigure}[b]{0.4\linewidth}
    \includegraphics[trim=0.5cm 0cm 4cm 2cm,clip,width=\linewidth]{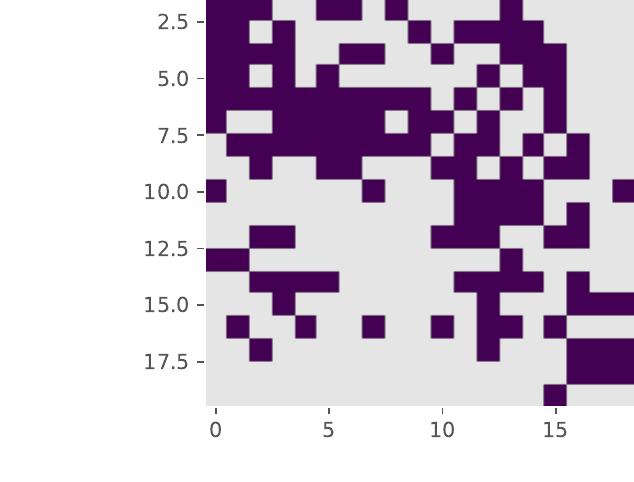}
    \caption{}
    \label{fig:graph3}
  \end{subfigure}\quad
  \begin{subfigure}[b]{0.4\linewidth}
    \includegraphics[trim=0.5cm 0cm 4cm 2cm,clip,width=\linewidth]{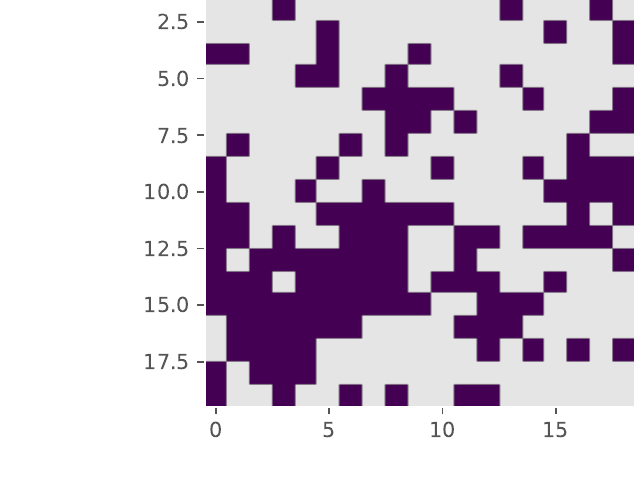}
    \caption{}
    \label{fig:graph4}
  \end{subfigure}
  \caption{The spatial distribution of the subsampling patterns. 
    a) Subsampling set $\mathcal{S}_1$. 
    b) Subsampling set $\mathcal{S}_2$.  
    c) Subsampling set $\mathcal{S}_3$.  
    d) Subsampling set $\mathcal{S}_4$.}
  \label{fig:graph345}
\end{figure}

Overall, the considered datasets cover a broad range of physical processes, spatial sampling regimes, and real-world variability, including purely advective dataset, advection–diffusion dynamics, large-scale atmospheric variables such as cloud cover and sea surface temperature, and real photovoltaic power generation. This experimental design enables us to evaluate the robustness and stability of the proposed framework across heterogeneous domains and underlying physical phenomena. Therefore, rather than focusing on performance gains in a single benchmark setting, our objective is to assess whether the proposed method consistently achieves competitive performance across diverse scenarios. We consider this cross-domain validation to provide a more informative assessment of generalization capability and practical relevance than repeated experiments on a single dataset. 

\subsection{Baselines}

 To clearly distinguish the different evaluation objectives, we organize the benchmarks as follows.  Two methods were used as benchmarks for forecasting evaluation across all datasets due to their ability to handle both regular and irregular grid predictions. The first benchmark is a graph convolutional long-short memory network (GCLSTM); as described in \cite{simeunovic2021spatio}. The GCLSTM framework consists of an encoder that ingests past signals data and then feeds its latent state to the decoder that made the final predictions. The combination of graph convolutional network and recurrent LSTM network is widely used in spatio-temporal forecasting tasks in different domains \citep{PAN2025132572, 9733962}. The second benchmark used is a simple graph convolutional LSTM model that has a single MLP after the LSTM layer (GCLSTMmlp). The combination of convolutional and recurrent structures (LSTM) has been used in many domains for time-series forecasting tasks \citep{shi2024time, shi2015convolutional, zhu2023drcnn}.

For datasets defined on regular grids, we additionally consider the following benchmark.  Additionaly, the FlowCNN model, a physics-informed approach proposed in \cite{de2019deep}, is designed specifically for regular grids and is a well-recognized approach for learning advection dynamics. However, its convolutional structure prevents its direct application to irregular sensor networks, which is why it was used only for datasets on regular grids, for sea surface temperature and cloud concentration index datasets.

In addition to forecasting performance, we evaluate the estimated transport dynamics.  We included optical flow as a baseline for velocity estimation, because it is a widely used and well-established method for estimating motion in image sequences. Our goal was to show that the velocity estimates produced by our model follow physical transport laws more closely than a naive optical flow estimate, and provides meaningful improvements that benefit the main forecasting objective. However, the accuracy of velocity estimation is not our primary goal and thus it was compared only with optical flow.

Finally, for the photovoltaic dataset, we compare the proposed model with a commercial forecasting approach.  Finally, the performance of the PING model integrated with GCLSTM was compared against a standard commercial solution \citep{satelliteremund} for PV power forecasting, which relies on cloud tracking models.

\subsection{Evaluation and metrics}
The model performance was evaluated using the peak normalized root-mean-square error (NRMSE), which is defined as:
\begin{equation}\label{eq:nrmse}
NRMSE(i, \tau) = \sqrt{\frac{1}{T} \sum_{t \in S} \left(\frac{\hat C_{i}^{t+\tau}-C_{i}^{t+\tau}}{{C^{max}_i}}\right)^2},
\end{equation}
where $C_i^{t+\tau}$ and  $\hat{C}_i^{t+\tau}$ represent the ground truth and predicted values, respectively, of node $i$ at time $t$ and horizon $\tau$. $C^{max}_i$ is the maximum value of node $i$ over the evaluation period $S$, and $T$ is the number of time steps in the evaluation interval $S$. For the PV power dataset, nighttime data is excluded from the error calculations.  

A second metric assesses the model's ability to estimates the velocity direction, by calculating the angle $\phi_{i}^t$ between the actual  $\mathbf v_i^t$ and estimated vector values $\mathbf {\hat v}_i^t$ at each time step $t$,  $t=1,...,6$: 
\begin{equation}\label{eq:phi}
\phi_{i}^t= \arccos \left( \frac{\mathbf v^t_i \cdot \mathbf {\hat v}_i^t}{|\mathbf v_i^t| |\mathbf {\hat v}_i^t|} \right),
\end{equation}
where $|\bold v_i^t|$ denotes the magnitude of the vector $\bold v_i^t$. This angle is calculated for every time step in the dataset and across all nodes $i$.

\subsection{Benchmarks}
 The evaluation is organized to distinguish between forecasting performance, velocity estimation, and comparison with commercial PV forecasting approach. In addition, we explicitly separate experiments on regular and irregular grids, since benchmark model FlowCNN is applicable only to regular domains. FlowCNN was selected as an additional benchmark because it is a well-established physics-aware deep learning approach for modeling advection phenomena. It represents one of the closest prior works to the proposed PING model in terms of incorporating physical inductive biases into the learning process. The performance and efficiency of the proposed model are first evaluated on synthetic and then on real-world datasets. 

In synthetic datasets on regular grid, where ground-truth velocity fields are available, the focus is on evaluating the model’s ability to accurately capture flow dynamics by aligning estimated velocity directions with the ground truth. In contrast, for real-world datasets such as cloud concentration index and sea surface temperature, ground-truth velocity measurements are not available, and therefore a direct evaluation of the velocity estimation is not possible.

 \subsubsection{Forecasting performance on regular grids}
We first evaluate the forecasting performance of the proposed model on regular grids.  In this setting, we compare PING with GCLSTM and GCLSTMmlp across all datasets, except for PV power forecasting, which is discussed separately.  For regular-grid datasets, we additionally include FlowCNN as a benchmark, since it is a well-established physics-aware model for learning advection dynamics but cannot be applied to irregular spatial domains.  Table \ref{tab:nrmse_reg_syn} presents the proposed model compared to GCLSTM and GCLSTMmlp on the synthetic advective-diffusive datasets evaluated on the regular grid, $22 \times 22$ windows. For the second half of the prediction horizon, from three to six steps ahead forecast, PING achieves higher accuracy than GCLSTM and GCLSTMmlp. On the purely advective dataset, PING outperforms GCLSTMmlp on the same part of the horizon and surpasses GCLSTM in the last three prediction steps. However, on this dataset, PING outperforms GCLSTM in the last three steps of the forecasting horizon. While PING and GCLSTMmlp use only MLP in the decoder, GCLSTM benefits from a recurrent network in the decoder, which propagates future dynamics.  These results highlight that PING gains advantage as the horizon extends on the regular grid. This is due to the design, which, by extracting spatial and temporal derivatives and incorporating attention, emphasizes the governing physical dynamics, which become more relevant as the prediction horizon increases. On the other hand, the recurrent structures in the decoder of GCLSTM are particularly effective at short-term state propagation.  

\begin{table}[h!]
\begin{center}
 \setlength\tabcolsep{5pt}
\caption{NRMSE of advection and advection-diffusion datasets on regular grid}
\small
\begin{tabular}{|cc|c|c|c|c|c|c|}
\hline
\multicolumn{2}{|c|}{Steps ahead}                                                                                              & \multirow{2}{*}{1} & \multirow{2}{*}{2} & \multirow{2}{*}{3} & \multirow{2}{*}{4} & \multirow{2}{*}{5} & \multirow{2}{*}{6} \\ \cline{1-2}
\multicolumn{1}{|c|}{dataset}                                                                                      & model     &                    &                    &                    &                    &                    &                    \\ \hline
\multicolumn{1}{|c|}{\multirow{3}{*}{\begin{tabular}[c]{@{}c@{}}Advection-\\ diffusion   \\  dataset\end{tabular}}}                & GCLSTM& \textbf{0.44}      & \textbf{0.87}      & \textbf{1.46}      & 2.19               & 3.08               & 4.28               \\ \cline{2-8} 
\multicolumn{1}{|c|}{}                                                                                             & GCLSTMmlp& 0.79               & 1.18               & 2.02               & 2.97               & 3.97               & 4.86               \\ \cline{2-8} 
\multicolumn{1}{|c|}{}                                                                                             & PING& 1.37               & 1.51               & 1.85               & \textbf{2.37}      & \textbf{2.9}       & \textbf{3.4}       \\ \hline
\multicolumn{1}{|c|}{\multirow{3}{*}{\begin{tabular}[c]{@{}c@{}}Advection\\ dataset\end{tabular}}} & GCLSTM& \textbf{1.51}      & \textbf{2.55}      & 3.69               & 4.87               & 6.18               & 7.73               \\ \cline{2-8} 
\multicolumn{1}{|c|}{}                                                                                             & GCLSTMmlp& 1.67               & 2.51               & 3.67               & 4.79               & 6.09               & 7.28               \\ \cline{2-8} 
\multicolumn{1}{|c|}{}                                                                                             & PING & 2.49               & 3.17               & \textbf{3.59}      & \textbf{3.99}      & \textbf{4.66}      & \textbf{5.85}      \\ \hline
\end{tabular}
\label{tab:nrmse_reg_syn}
\end{center}
\end{table}

\begin{table}[h]
\begin{center}
 \setlength\tabcolsep{5pt}
\caption{NRMSE of cloud concentration index and sea surface temperature datasets on regular grid}
\small
\begin{tabular}{|cc|c|c|c|c|c|c|}
\hline
\multicolumn{2}{|c|}{Steps ahead} & \multirow{2}{*}{1} & \multirow{2}{*}{2} & \multirow{2}{*}{3} & \multirow{2}{*}{4} & \multirow{2}{*}{5} & \multirow{2}{*}{6} \\ \cline{1-2}
\multicolumn{1}{|c|}{dataset} & model &  &  &  &  &  &  \\ \hline
\multicolumn{1}{|c|}{\multirow{4}{*}{\begin{tabular}[c]{@{}c@{}}Cloud \\ index  \\  dataset\end{tabular}}} & GCLSTM & 12.83 & 18.56 & {22.03} & 24.27 & 25.97 & 27.26 \\ \cline{2-8} 
\multicolumn{1}{|c|}{} & GCLSTMmlp & 13.39 & 19.78 & 23.24 & 25.44 & 26.9 & 28.07 \\ \cline{2-8} 
\multicolumn{1}{|c|}{} & FlowCNN & 13.49 & 18.55 & 21.58 & 23.71 & 25.3 & 26.61 \\ \cline{2-8} 
\multicolumn{1}{|c|}{} & PING & \textbf{12.21} & \textbf{17.42} & \textbf{20.93} & \textbf{23.47} & \textbf{25.29} & \textbf{26.57} \\ \hline
\multicolumn{1}{|c|}{\multirow{4}{*}{\begin{tabular}[c]{@{}c@{}}SST \\ dataset\end{tabular}}} & GCLSTM & \textbf{1.67} & \textbf{2.29} & 2.89 & 3.47 & 4.01 & 4.48 \\ \cline{2-8} 
\multicolumn{1}{|c|}{} & GCLSTMmlp & 1.69 & 2.31 & 2.93 & 3.5 & 4.02 & 4.45 \\ \cline{2-8} 
\multicolumn{1}{|c|}{} & FlowCNN & 1.69 & 2.39 & 3.06 & 3.67 & 4.26 & 4.8 \\ \cline{2-8} 
\multicolumn{1}{|c|}{} & PING & 1.81 & 2.34 & \textbf{2.87} & \textbf{3.38} & \textbf{3.85} & \textbf{4.26} \\ \hline
\end{tabular}
\label{tab:nrmse_reg_weather}
\end{center}
\end{table}

In Table \ref{tab:nrmse_reg_weather}, the performance of the PING model is compared to the GCLSTM, FlowCNN and GCLSTMmlp on a regular grid for the cloud concentration and sea surface temperature datasets from ERA5. The PING model outperforms GCLSTM across the entire horizon on the cloud dataset and surpasses FlowCNN for the first five hours of predictions. For smoother diffusive processes, such as sea surface temperature, PING performs comparably to FlowCNN for the first horizon steps and exceeds FlowCNN’s accuracy from the second to sixth days of predictions. On the same dataset, PING outperforms GCLSTM from the third to sixth steps, where global dynamics start influencing concentration estimates.  This difference between the cloud and sea surface temperature datasets illustrates PING’s advantage in capturing advective dynamics. In smoother, more diffusive processes, PING’s advantage appears later in the prediction horizon, when large-scale dynamics become more important.

\subsubsection{Forecasting performance on irregular grids}

We next evaluate forecasting performance on irregular spatial domains, where only graph-based models can be applied. In this setting, FlowCNN is not included, and the comparison is conducted between PING, GCLSTM and GCLSTMmlp. 

The results for irregularly subsampled domains with 150 points are shown in Table \ref{tab:nrmse_ireg_syn}. The PING model performs similarly on irregular domains as on the regular grid. It surpasses GCLSTM and GCLSTMmlp in accuracy from the third to the sixth steps ahead on advective-diffusive datasets. For the purely advective dataset, PING outperforms GCLSTMmlp in the same part of the horizon and exceeds GCLSTM's accuracy for the last two steps of the forecast horizon. Across these datasets, the PING model achieves a median NRMSE of 2\% to 4\%  for the advective-diffusive datasets and 3\% to 6.5\% for the purely advective dataset, indicating limited sensitivity of accuracy to sampling patterns. There is a slight difference in the accuracy between irregularly sampled advection datasets. However, this is due to the fact that the direct source of the smoke in PhiFLow simulations is in the left-bottom quadrant and predicting it is extremely difficult. Hence, $\mathcal{S}_1$ shows the highest error because it contains the most pixels in that region, while $\mathcal{S}_3$ achieves the highest accuracy because it has the fewest pixels in that quadrant. Although the absolute performance gap between the PING and baseline models is small, it suggests that PING’s embedding  and attention modules help maintain accuracy under irregular sampling while finding meaningful patterns between nodes, while the baselines show somewhat higher sensitivity to sampling geometry. The ability to generalize across sampling patterns is especially important for real-world observational datasets where spatial coverage is non-uniform. 

\begin{table}[h]
\begin{center}
 \setlength\tabcolsep{5pt}
\caption{NRMSE of advection datasets on graph instances $S_1, S_2, S_3, S_4$}
\small
\begin{tabular}{|ll|l|l|l|l|l|l|}
\hline
\multicolumn{2}{|l|}{Steps ahead} & \multirow{2}{*}{1} & \multirow{2}{*}{2} & \multirow{2}{*}{3} & \multirow{2}{*}{4} & \multirow{2}{*}{5} & \multirow{2}{*}{6} \\ \cline{1-2}
\multicolumn{1}{|l|}{model} & \begin{tabular}[c]{@{}l@{}}Graph\\  instance\end{tabular} &  &  &  &  &  &  \\ \hline
\multicolumn{1}{|l|}{GCLSTM} & \multirow{3}{*}{S1} & \textbf{1.96} & \textbf{2.73} & \textbf{4.06} & 5.62 & 7.18 & 8.42 \\ \cline{1-1}  \cline{3-8} 
\multicolumn{1}{|l|}{GCLSTMmlp} &  & 1.96 & 3.1 & 4.19 & 5.83 & 7.38 & 8.64 \\ \cline{1-1}  \cline{3-8} 
\multicolumn{1}{|l|}{PING} &  & 3.11 & 3.97 & 4.65 & \textbf{5.61} & \textbf{6.74} & \textbf{7.76} \\ \hline
\multicolumn{1}{|l|}{GCLSTM} & \multirow{3}{*}{S2} & \textbf{1.92} & \textbf{3.22} & 4.69 & 5.9 & 7.23 & 8.48 \\ \cline{1-1}  \cline{3-8} 
\multicolumn{1}{|l|}{GCLSTMmlp} &  & 2.58 & 3.53 & 4.63 & 5.44 & 6.2 & 7.2 \\ \cline{1-1}  \cline{3-8} 
\multicolumn{1}{|l|}{PING} &  & 2.98 & 3.56 & \textbf{4.04} & \textbf{4.83} & \textbf{5.63} & \textbf{6.58} \\ \hline

\multicolumn{1}{|l|}{GCLSTM} & \multirow{3}{*}{S3} & \textbf{1.74} & \textbf{2.64} & 3.64 & 4.69 & 5.86 & 7.32 \\ \cline{1-1} \cline{3-8} 
\multicolumn{1}{|l|}{GCLSTMmlp} &  & 2.15 & 3.09 & 3.67 & 4.75 & 5.96 & 7.12 \\ \cline{1-1} \cline{3-8} 
\multicolumn{1}{|l|}{PING} &  & 2.49 & 3.18 & \textbf{3.59} & \textbf{3.99} &\textbf{ 4.66} & \textbf{5.48} \\ \hline
\multicolumn{1}{|l|}{GCLSTM} & \multirow{3}{*}{S4} &  1.75 &  2.53 &  \textbf{3.44} & 4.65 & 6.44 & 8.15 \\ \cline{1-1} \cline{3-8} 
\multicolumn{1}{|l|}{GCLSTMmlp} &  & \textbf{1.70} & \textbf{2.46} & 3.57 & 4.94 & 5.87 & 6.46 \\ \cline{1-1} \cline{3-8} 
\multicolumn{1}{|l|}{PING} &  & 2.79 & 3.36 & 3.87 & \textbf{ 4.64} &  \textbf{5.23} &  \textbf{6.27} \\ \hline
\end{tabular}
\label{tab:nrmse_ireg_syn}
\end{center}
\end{table}

Finally, Tables \ref{tab:nrmse_ireg_weather} and \ref{tab:nrmse_ireg_sst} summarize the model's performance on irregularly sampled points for each real dataset.  Interestingly, on the SST and cloud concentration index datasets, the forecasting accuracy is the same across all four types of subsampling.  The PING model consistently outperforms GCLSTM across the entire horizon for the cloud dataset on all irregular samples. For the sea surface temperature dataset, PING surpasses GCLSTM in accuracy from the third to sixth prediction steps.  These findings reveal a consistent pattern that PING has a clear advantage as the horizon extends across both synthetic and real datasets under regular and irregular sampling. This underscores that physics-informed architectures can provide longer-term robustness compared to purely recurrent and convolutional baselines that excel at short horizons. 

\begin{table}[h]
\begin{center}
 \setlength\tabcolsep{5pt}
\caption{NRMSE of  cloud concentration index datasets on irregular grid}
\small
\begin{tabular}{|ll|l|l|l|l|l|l|}
\hline
\multicolumn{2}{|l|}{Steps ahead} & \multirow{2}{*}{1} & \multirow{2}{*}{2} & \multirow{2}{*}{3} & \multirow{2}{*}{4} & \multirow{2}{*}{5} & \multirow{2}{*}{6} \\ \cline{1-2}
\multicolumn{1}{|l|}{model} & \begin{tabular}[c]{@{}l@{}}Graph \\ instance\end{tabular} &  &  &  &  &  &  \\ \hline
\multicolumn{1}{|l|}{GCLSTM} & \multirow{3}{*}{S1} &  \textbf{12.52} & 18.07 & 21.29 & 23.46 & 25.07 & 26.34 \\ \cline{1-1}\cline{3-8} 
\multicolumn{1}{|l|}{GCLSTMmlp} &  & 13.12 & 19.1 & 22.57 & 24.69 & 26.22 & 27.38 \\ \cline{1-1} \cline{3-8} 
\multicolumn{1}{|l|}{PING} &  & 12.64 &  \textbf{17.56} &  \textbf{20.83} &  \textbf{23.13} &  \textbf{24.9} &  \textbf{26.12} \\ \cline{1-1} \cline{2-8} 

\multicolumn{1}{|l|}{GCLSTM} & \multirow{3}{*}{S2} & \textbf{12.55} & 18.01 & {21.35} & {23.58} & {25.24} & {26.66} \\ \cline{1-1} \cline{3-8} 
\multicolumn{1}{|l|}{GCLSTMmlp} &  & {16.45} & {19.52} & 23.12 & 25.38 & 26.86 & 27.9 \\ \cline{1-1} \cline{3-8} 
\multicolumn{1}{|l|}{PING} &  & 12.61 &  \textbf{17.52} & \textbf{ 20.82} &  \textbf{23.25} &  \textbf{24.98 }&  \textbf{26.32} \\ \cline{1-1} \cline{2-8} 

\multicolumn{1}{|l|}{GCLSTM} & \multirow{3}{*}{S3} & 12.62 & 18.18 & 21.49 & 23.78 & 25.44 & 26.84 \\ \cline{1-1} \cline{3-8} 
\multicolumn{1}{|l|}{GCLSTMmlp} &  & 13.26 & 19.29 & {22.77} & {25.08} & {26.57} & {27.66} \\ \cline{1-1} \cline{1-1} \cline{3-8} 
\multicolumn{1}{|l|}{PING} &  & \textbf{12.59} &  \textbf{17.41} &  \textbf{20.64} &  \textbf{22.99} & \textbf{ 24.77 }&  \textbf{26.05} \\ \hline

\multicolumn{1}{|l|}{GCLSTM} & \multirow{3}{*}{S4} & \textbf{12.55} & 17.98 & 21.31 & 12.51 & 25.18 & 26.65 \\ \cline{1-1} \cline{1-1} \cline{3-8} 
\multicolumn{1}{|l|}{GCLSTMmlp} &  & 13.48 & 19.64 & 22.91 & 25.06 & 26.47 & 27.57 \\ \cline{1-1} \cline{1-1} \cline{3-8} 
\multicolumn{1}{|l|}{PING} &  & 12.62 & \textbf{17.59} & \textbf{20.77} & \textbf{23.08} & \textbf{24.9} & \textbf{26.33} \\ \hline
\end{tabular}
\label{tab:nrmse_ireg_weather}
\end{center}
\end{table}

\begin{table}[h]
\begin{center}
 \setlength\tabcolsep{5pt}
\caption{NRMSE of sea surface temperature datasets on irregular grid}
\small
\begin{tabular}{|ll|l|l|l|l|l|l|}
\hline
\multicolumn{2}{|l|}{Steps ahead} & \multirow{2}{*}{1} & \multirow{2}{*}{2} & \multirow{2}{*}{3} & \multirow{2}{*}{4} & \multirow{2}{*}{5} & \multirow{2}{*}{6} \\ \cline{1-2}
\multicolumn{1}{|l|}{model} & \begin{tabular}[c]{@{}l@{}}Graph \\ instance\end{tabular} &  &  &  &  &  &  \\ \hline
\multicolumn{1}{|l|}{GCLSTM} & \multirow{3}{*}{S1} & \textbf{1.67} & \textbf{2.29} & {2.89} & 3.45 & 4.01 & 4.52 \\ \cline{1-1}\cline{3-8} 
\multicolumn{1}{|l|}{GCLSTMmlp} &  & 1.77 & 2.34 & 2.94 & 3.51 & 4.02 & 4.54 \\ \cline{1-1}\cline{3-8} 
\multicolumn{1}{|l|}{PING} &  & 1.88 & 2.44 & 2.92 & \textbf{3.37 }&\textbf{ 3.9 } & \textbf{4.31 }\\ \cline{1-1}\cline{2-8} 
\multicolumn{1}{|l|}{GCLSTM} & \multirow{3}{*}{S2} & \textbf{1.69} & \textbf{2.29} & \textbf{2.88} & \textbf{3.43} & {3.98} & {4.51} \\ \cline{1-1} \cline{3-8} 
\multicolumn{1}{|l|}{GCLSTMmlp} &  & {1.95} & {2.52} & 3.1 & 3.66 & 4.2 & 4.56 \\ \cline{1-1}\cline{3-8} 
\multicolumn{1}{|l|}{PING} &  & 2.14 & 2.57 & 3.05 & 3.51 & \textbf{3.91} & \textbf{4.37} \\ \cline{1-1} \cline{2-8} 
\multicolumn{1}{|l|}{GCLSTM} & \multirow{3}{*}{S3} & \textbf{1.72} & \textbf{2.31} &\textbf{ 2.91} & 3.47 & 4.02 & 4.52 \\ \cline{1-1}\cline{3-8} 
\multicolumn{1}{|l|}{GCLSTMmlp} &  & 1.88 & 2.48 & {3.08} & {3.65} & {4.18} & {4.63} \\ \cline{1-1} \cline{1-1} \cline{3-8} 
\multicolumn{1}{|l|}{PING} &  & 1.94 & 2.44 & 2.95 & \textbf{3.41} & \textbf{3.88} & \textbf{4.29} \\ \hline
\multicolumn{1}{|l|}{GCLSTM} & \multirow{3}{*}{S4} & \textbf{1.72} & \textbf{2.34} & 2.94 & 3.47 & 4.01 & 4.49 \\ \cline{1-1} \cline{3-8} 
\multicolumn{1}{|l|}{GCLSTMmlp} &  & 1.88 & 2.47 & 3.08 & 3.65 & 4.18 & 4.63 \\ \cline{1-1} \cline{3-8} 
\multicolumn{1}{|l|}{PING} &  & 1.86 & 2.42 & \textbf{2.92} &\textbf{ 3.44} & \textbf{3.91} & \textbf{4.32} \\ \hline
\end{tabular}
\label{tab:nrmse_ireg_sst}
\end{center}
\end{table}

\subsubsection{Performance at longer forecasting horizons}

Across both synthetic and real-world datasets, we notice a pattern that the proposed PING model has an advantage over baseline methods as the forecasting horizon extends. On synthetic datasets, this behavior is particularly evident for both advective and advection–diffusion dynamics, where PING achieves higher accuracy from the third to sixth prediction steps on both regular and irregular grids. This trend is also observed in real-world datasets. For the cloud concentration index, PING outperforms GCLSTM across the entire horizon, while for smoother and more diffusive processes such as sea surface temperature, the improvement becomes more pronounced at longer horizons, where global dynamics play main role.

This behavior can be attributed to the physics-informed design of the proposed framework. By incorporating spatial and temporal derivatives and the dynamics, PING captures the underlying physical processes that govern long-term evolution. In contrast, purely data-driven recurrent and convolutional baselines are more effective at short-term state propagation but tend to accumulate errors over longer horizons. The results on both regular and irregular spatial domains indicate that physics-informed modeling improves robustness and stability in long-term forecasting, particularly under irregular sampling and diverse physical conditions.

\subsubsection{Sensitivity to regularization parameters}

We analyze the sensitivity of the proposed framework to the physics-informed loss weights, where $\lambda_1$ controls the advection--diffusion residual and $\lambda_2$ enforces the incompressibility constraint. This study is conducted on the advection--diffusion dataset under irregular sampling.

Table~\ref{tab:lambda_sensitivity} summarizes the forecasting accuracy for different values of $\lambda_1$ and $\lambda_2$. The results indicate that increasing $\lambda_1$ from 0.15 to 0.20 while keeping $\lambda_2=0.05$ improves forecasting accuracy across all prediction horizons. This suggests that strengthening the advection--diffusion constraint helps stabilize long-term predictions.

In contrast, increasing the incompressibility weight $\lambda_2$ to 0.1 does not improve the results and slightly degrades forecasting accuracy, particularly at longer horizons. This behavior indicates that overly strong divergence-free regularization may overconstrain the estimated velocity field and reduce model flexibility.

These observations provide qualitative insight into the role of the physics-informed regularization terms. Based on this study, we use $(\lambda_1,\lambda_2) = (0.2, 0.05)$ in all subsequent experiments.

\begin{table}[ht]

\centering
\caption{ NRMSE sensitivity analysis of the physics-informed loss weights on the advection--diffusion dataset under irregular sampling.}
\small
\setlength{\tabcolsep}{6pt}
\renewcommand{\arraystretch}{1.3}
\begin{tabular}{|c|c|c|c|c|c|c|}
\hline
\multicolumn{7}{|c|}{Steps ahead} \\ \hline
 & 1 & 2 & 3 & 4 & 5 & 6 \\ \hline
$\lambda_1=0.15,\ \lambda_2=0.05$ 
& 0.0321 & 0.0389 & 0.0404 & 0.0480 & 0.0548 & 0.0619 \\ \hline

$\lambda_1=0.20,\ \lambda_2=0.05$ 
& 0.0250 & 0.0318 & 0.0360 & 0.0399 & 0.0466 & 0.0548 \\ \hline

$\lambda_1=0.15,\ \lambda_2=0.10$ 
& 0.0299 & 0.0352 & 0.0430 & 0.0478 & 0.0574 & 0.0666 \\ \hline
\end{tabular}
\label{tab:lambda_sensitivity}
\end{table}

\subsubsection{Velocity direction analysis}
In addition to forecasting performance, we evaluate the quality of the estimated transport dynamics.  Equation \ref{eq:phi} is used to evaluate whether the velocity field reconstructed by the flow estimator, described in Section \ref{flow_estim}, aligns with the observed velocities. The results are compared against optical flow estimates obtained using the Gunnar-Farneback model   \citep{farneback2003two}.  Table \ref{tab:velocity_optical} demonstrates that PING  outperforms Gunnar-Farneback optical flow method on a regular grid for both synthetic datasets. 

\begin{table}[ht]
\centering
\caption{Accuracy of velocity estimation on regular grid for advection and advection-diffusion datasets (PING vs. Optical flows)}
\small
\renewcommand{\arraystretch}{1.5}  
\setlength{\tabcolsep}{10pt}       
\begin{tabular}{|l|l|l|l|l|}
\hline
\textbf{Dataset}                            & \textbf{Model} & \textbf{$\phi(rad)$}  \\ \hline
\multirow{2}{*}{Advection dataset}          & PING           & \bf{0.44}                               \\ \cline{2-3} 
                                            & Optical flow   & 1.16                     \\ \hline
\multirow{2}{*}{Advection-diffusion dataset}& PING           & \bf{0.92}          \\ \cline{2-3} 
                                            & Optical flow   & 1.10             \\ \hline
\end{tabular}
 \label{tab:velocity_optical}
\end{table}

\begin{figure}[h!] 
 \centering 
    \includegraphics[trim=2cm 2cm 2cm 2.0cm, clip, width=0.6\linewidth]{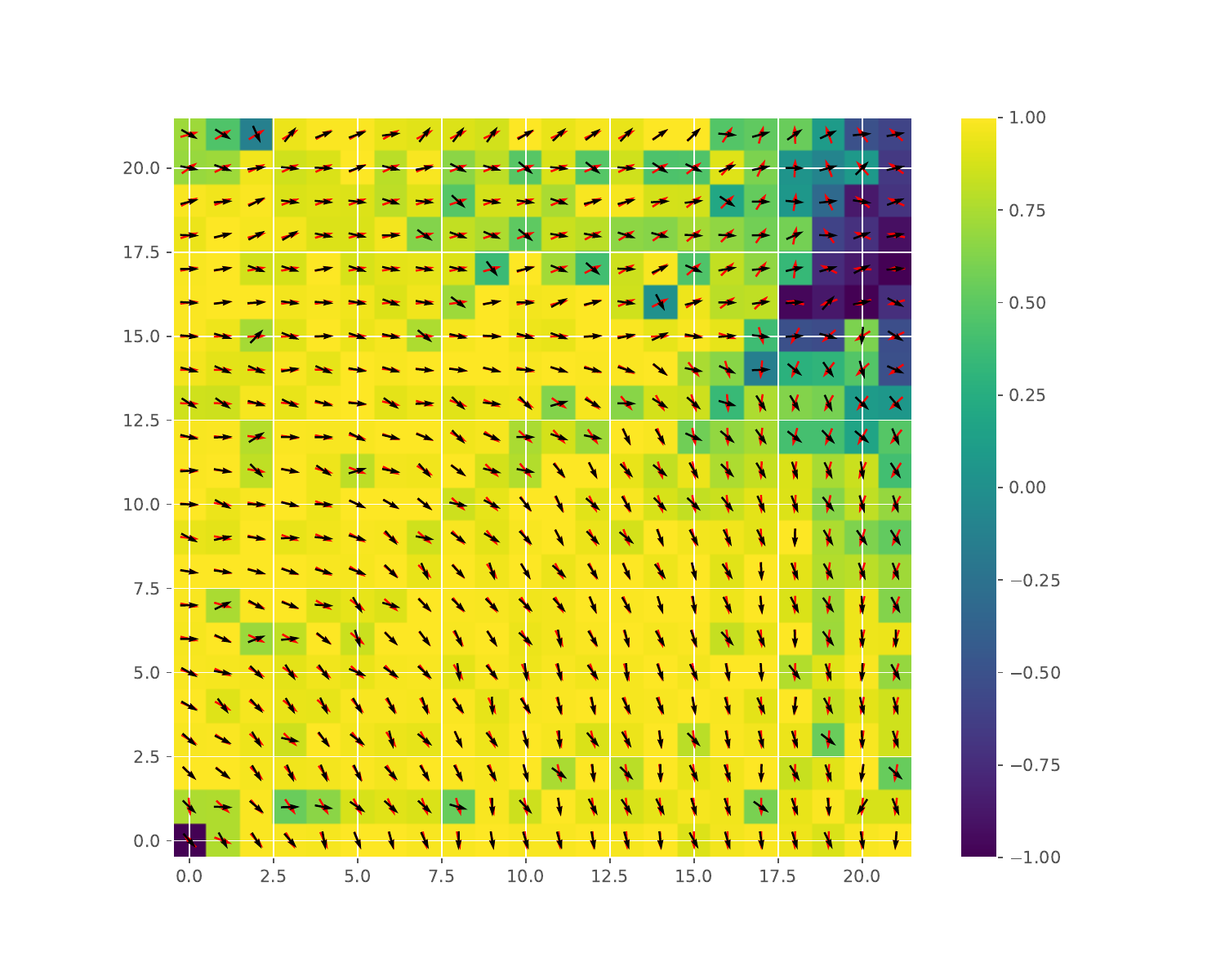}
              \caption{Example of velocities estimation on a regular grid, for the advection only dataset. The similarity between the ground truth velocity direction and the estimated one is color-coded: yellow represents the perfect alignment between the vectors and darker colors indicate larger deviations. The arrows represent the direction of the velocity vector, black being the ground truth and red the estimated one.} 
              \label{fig:adv_higlight_reg}
              \end{figure}

\begin{figure}[h!] 
 \centering 
    \includegraphics[trim=2cm 2cm 2cm 2.0cm, clip, width=0.6\linewidth]{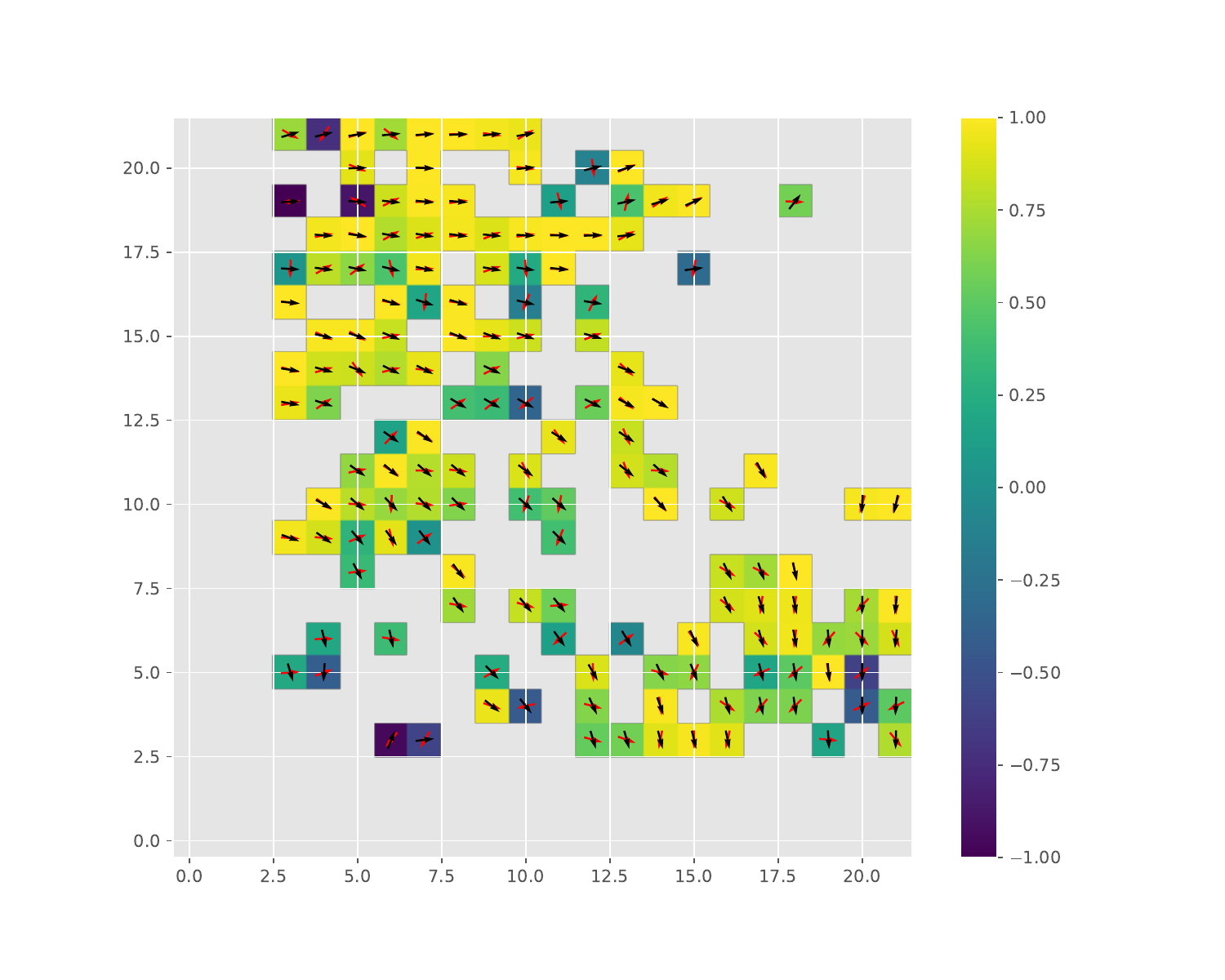}
      \caption{Example of velocities estimation on an irregular grid, for the advection only dataset. The similarity between the ground truth velocity direction and the estimated one is color-coded: yellow represents the perfect alignment between the vectors and darker colors indicate larger deviations. The arrows represent the direction of the velocity vector, black being the ground truth and red the estimated one.} 
              \label{fig:adv_higlight_ireg}
              \end{figure}

Figures  \ref{fig:adv_higlight_reg} and \ref{fig:adv_higlight_ireg} illustrate the velocity estimates compared to the ground truth for both regular and irregular grids. A color-coding scheme is applied to each node to visualize the alignment between the estimated and observed velocity vectors. The black arrows represent the ground truth velocity directions, while red arrows denote the estimated flow directions. Yellow shades of the node represent the scenario where two vectors are coincident, green shades indicate the situation when vectors are orthogonal, and finally, dark blue represents vectors with opposite directions. Figure \ref{fig:adv_higlight_reg} highlights the model's ability to accurately estimate the flow direction when the concentration changes are smooth on the regular grid. However, on the irregular grids, the task is more challenging due to partial information from the sampled nodes.

\subsubsection{Application to PV power forecasting}
\begin{figure}[h!]
  \centering
    \includegraphics[trim=1cm 4.5cm 3.2cm 6cm, clip, width=0.9\linewidth]{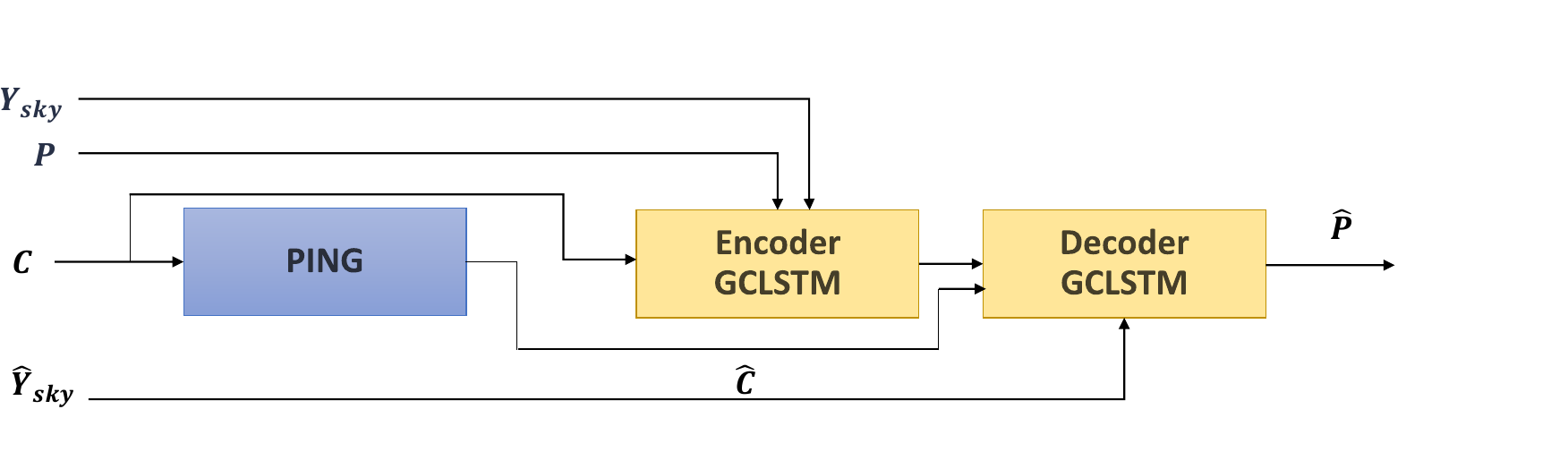}
  \caption{Combination of PING and GCLSTM for PV power forecasting. PING takes as inputs past cloud cover index values $C$. The encoder of GCLSTM takes as inputs cloud cover index $\bf C$ values, historical clear-sky values $\bf Y$ and historical observations from  PV power plants $\bf P$. The output from PING, the encoder state and future clear-sky values are passed to the decoder.  }
\label{fig:nrmse_ping_gclstm_model}
\end{figure}

Finally, we assess the practical relevance of the proposed approach by comparing the PING model integrated with GCLSTM with a standard commercial PV forecasting solution based on cloud tracking.  The results presented demonstrate that PING outperforms other models on the cloud concentration index dataset for both regular and irregular grids. In this section, the PING model is utilized to forecast the cloud index concentration and is integrated into a previously developed PV power forecasting framework based on GCLSTM. The proposed combination of PING and GCLSTM is illustrated in Figure \ref{fig:nrmse_ping_gclstm_model}. The PING model estimates the cloud motion using historical cloud concentration index values $\bf{C}$ and predicts future values $\hat{\bf{C}}$. The past cloud concentration values $\bf{C}$ are concatenated with the past production data $\bf{P}$ and clear-sky irradiance $\bf{Y_{sky}}$, serving a inputs to the encoder of the GCLSTM model. The GCLSTM uses encoder state along with the predicted future cloud concentration values $\hat{\bf{C}}$ from PING model and future clear-sky irradiance values $\bf{\hat{Y}_{sky}}$. Clear-sky irradiance is a deterministic variable that can be calculated for any time of the year at a specific location. 
 This setup allows PING to specialize in learning cloud motion dynamics, while GCLSTM integrates this information with ground-level PV and irradiance data. We evaluate 6-hour-ahead forecasts at 15-minute resolution, since this horizon is most relevant for grid congestion management and energy trading. Shorter horizons are typically addressed by persistence or cloud-tracking solutions based on all-sky imagers, while longer horizons are better suited for numerical weather prediction models.

\begin{figure}[h!]
  \centering
    \includegraphics[trim=1cm 0.5cm 1cm 1cm, clip, width=0.6\linewidth]{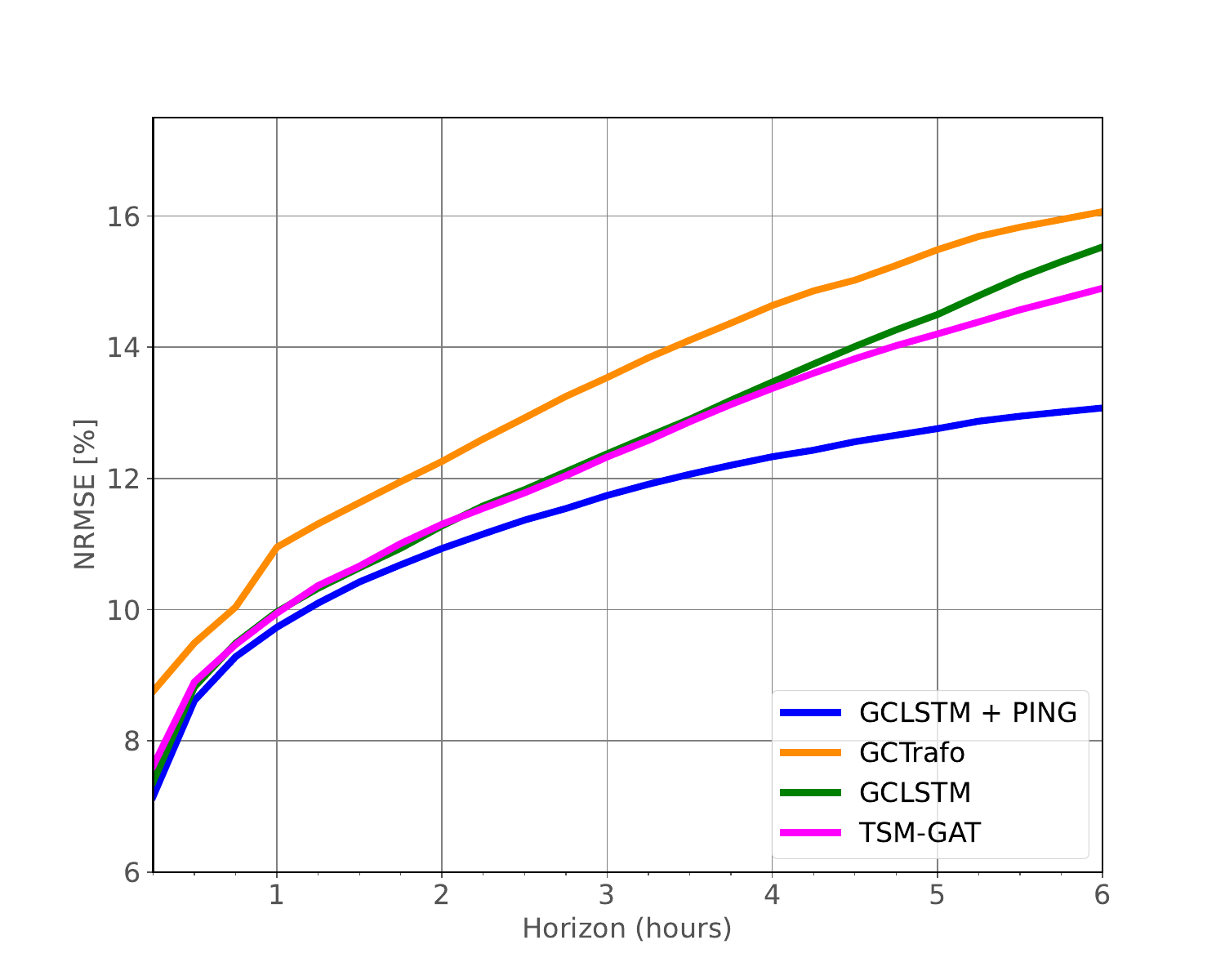}
  \caption{Evolution of the NRMSE between GCLSTM+PING, GCLSTMmlp and GCLSTM models for six hours ahead for PV power generation, with hourly resolution. The solid line shows the median value among all nodes.}
\label{fig:nrmse_pv}
\end{figure}

The combination of PING and GCLSTM is evaluated on a PV power generation dataset and compared against several benchmarks that do not use satellite data, only the power measured at the ground PV stations, including GCLSTM, GCTrafo   \citep{simeunovic2021spatio},  and the TSM-GAT model   \citep{simeunovic2022interpretable}.   These models were chosen since they are based on convolutional networks, recurrent networks, transformers and graph attention networks and cover a variety of spatio-temporal forecasting architectures for time-series forecasting tasks. They have shown state-of-the-art results for six-hour-ahead PV power forecasting.  As shown in Figure \ref{fig:nrmse_pv}, the combination of PING+GCLSTM outperforms all benchmarks on the entire horizon across a six-hour forecasting horizon at a resolution of fifteen minutes.  This indicates that modeling cloud dynamics with the  PING model adds useful information  to  the PV power forecasting task. The benefits are more visible at longer horizons, when cloud dynamics   become  more relevant than  the  autoregressive component. 

\begin{figure}[h!]
\centering
\includegraphics[trim=0cm 0cm 0.0cm 0cm, clip, width=0.65\linewidth]{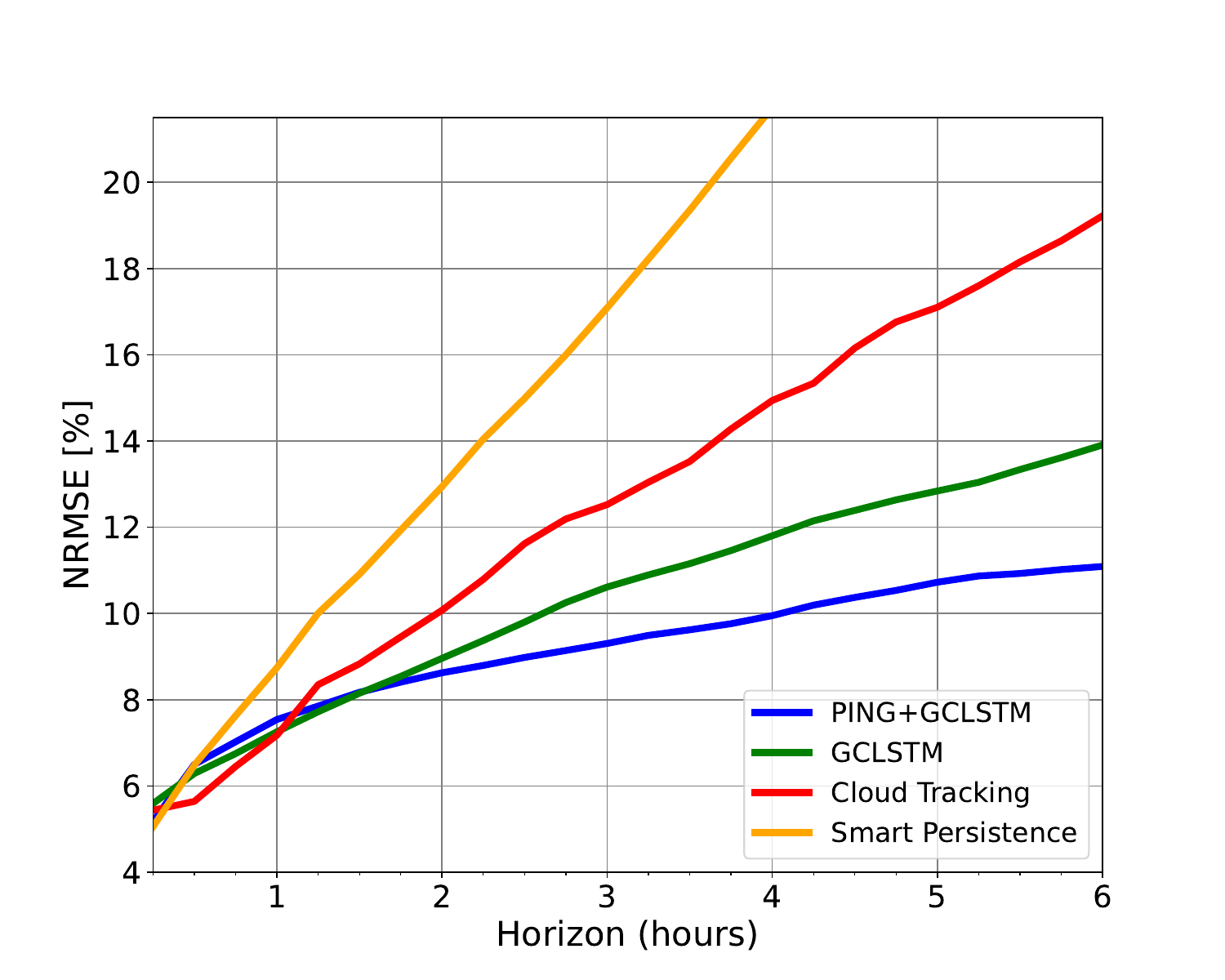}
\caption{NRMSE evolution of GCLSTM, CloudMove, GCLSTM+PING and persistence model over the forecasting horizon of 6 hours ahead in steps of 15 minutes. }
\label{fig:clooud_tracking}
\end{figure}

Additionally, the GCLSTM+PING model was compared with the cloud tracking commercial solution, CloudMove, under different seasons and weather conditions. The comparison to CloudMove ensures fair comparison to both research and industry standards. A representative set of 18 locations and 21 days were selected over the year 2017 to cover the whole range of possible weather conditions, terrain types, and distances between irradiance measurement stations in Switzerland. Details on the selected locations and dates can be found in \cite{clc_rcr}. Figure \ref{fig:clooud_tracking} shows the NRMSE over the six-hour forecast horizon in 15-minute intervals. While GCLSTM+PING exhibits a larger error spread than CloudMove, it achieves lower NRMSE after one and a half hours and outperforms CloudMove with an 8\% lower NRMSE in six-hour forecasts. Compared to GCLSTM alone, the combined model reduces errors by 3\%.  This indicates that commercial methods such as CloudMove may provide competitive very-short-term forecasts, but physics-informed graph modeling becomes advantageous for forecasts beyond 2 hours. The higher variance of GCLSTM+PING errors suggests that it is more sensitive to rapid changes in the cloud field, while its lower median error highlights robustness for longer forecasting horizons.

\section{Conclusion}
\label{section_5}
We developed a novel model, PING, designed to forecast scalar quantities involved in transport phenomena and historical past velocities in an unsupervised manner  on irregularly sampled grids. The model was evaluated on a diverse set of datasets, including sea surface temperature and cloud concentration index from ERA5, as well as two synthetic datasets simulating advection and advection-diffusion processes. Its performance was benchmarked against baseline models on the previously mentioned datasets under both regular and irregular grids. Across these experiments, PING achieved accuracy on par with state-of-the-art baselines for sea surface temperature and synthetic datasets, while showing clear improvements in the cloud datasets across the entire forecast horizon. Importantly, PING demonstrated robustness to irregular sampling.

Beyond standalone evaluation, we demonstrated the practical value of PING in PV power forecasting by integrating it with GCLSTM. In this hybrid framework, PING’s forecasts of future cloud concentration were used as input to the GCLSTM decoder, allowing explicit modeling of cloud motion in power predictions. This combined approach outperformed state-of-the-art learning-based baselines (GCLSTM, GCTrafo, TSM-GAT) and a commercial cloud-tracking solution (CloudMove), reducing forecasting errors by 3\% in NRMSE compared to GCLSTM alone and by 8\% compared to CloudMove over a six-hour horizon.

While these results demonstrate the strengths of PING, several limitations remain. First, in discretizing the governing equations, we assumed purely two-dimensional horizontal dynamics. This approximation becomes inaccurate in multi-layered flows and needs to be further addressed. Although PING consistently outperforms baselines at longer horizons, its short-term performance is sometimes comparable, highlighting that recurrent decoders still have an advantage in propagating immediate dynamics. Finally, the regularization coefficients in the physics-informed loss were chosen empirically and a more systematic study of their sensitivity could provide further insight.

Extending PING to a three-dimensional or multi-layer representation using tensor-based discretizations would be a promising direction for future work, as it would broaden the model's applicability. Additionally, increasing the forecasting horizon for day-ahead PV power forecasts would be another promising direction for future work, since PING showed promising results at longer parts of the forecasting horizon.

\section*{Acknowledgments}
This work received the funding from CSEM Data Program. The authors are solely responsible for the content and conclusions of the paper.
 
\appendix
\section{Hyperparameters}
For all models, we performed hyperparameter tuning using manual tuning. The training was done with stochastic gradient descent and the Adam optimizer with a learning rate 1e-4. The baseline models GCRNN and GCRNNmlp were trained up to 200.000 iterations, while the PING was trained up to 700.000 iterations. For the most datasets, the GCRNN and GCRNNmlp were stopped early before 80.000 iterations, while PING was early stopped at 500.000 iterations. The difference in the number of training iterations comes from higher number of learnable weights in the PING model compared to GCRNN or GCRNNmlp. The model is trained using the mean square error (MSE) loss function and early stopping was applied. All models, including baselines, were trained using the same data splits and under the same computational setup to ensure a fair comparison. All graph-based models are using graphs with 8 nodes as nearest neighbors. The models were trained and evaluated on a workstation with 64 cores, 503 GB of RAM memory and an Nvidia  GeForce RTX 3090 GPU.

\section{Evaluation analysis of predicted signals}
Further illustrative analysis compares forecasts made at different times of the day and at particular spatial nodes. These examples are provided to showcase how the proposed model behaves under varying conditions and are not intended as comprehensive evaluations.

\begin{figure}[h!]
    \begin{subfigure}{0.5\linewidth}
    \includegraphics[trim=0.2cm 0.8cm 2cm 2.0cm, clip, width=\linewidth]{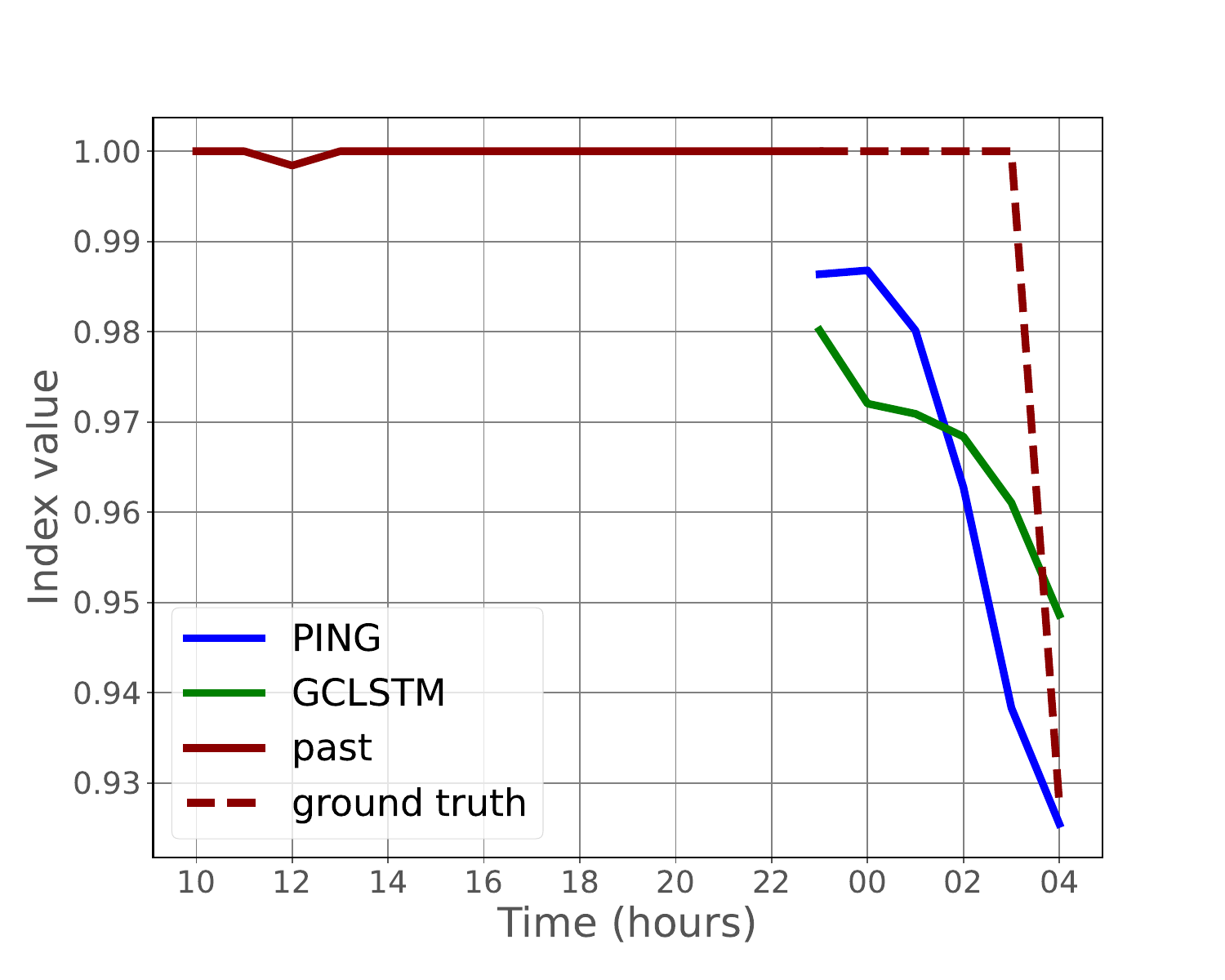}
     \label{fig:reg11}
  \end{subfigure} 
\hfill
    \begin{subfigure}{0.5\linewidth}
    \includegraphics[trim=0.2cm 0.8cm 2cm 2.0cm, clip, width=\linewidth]{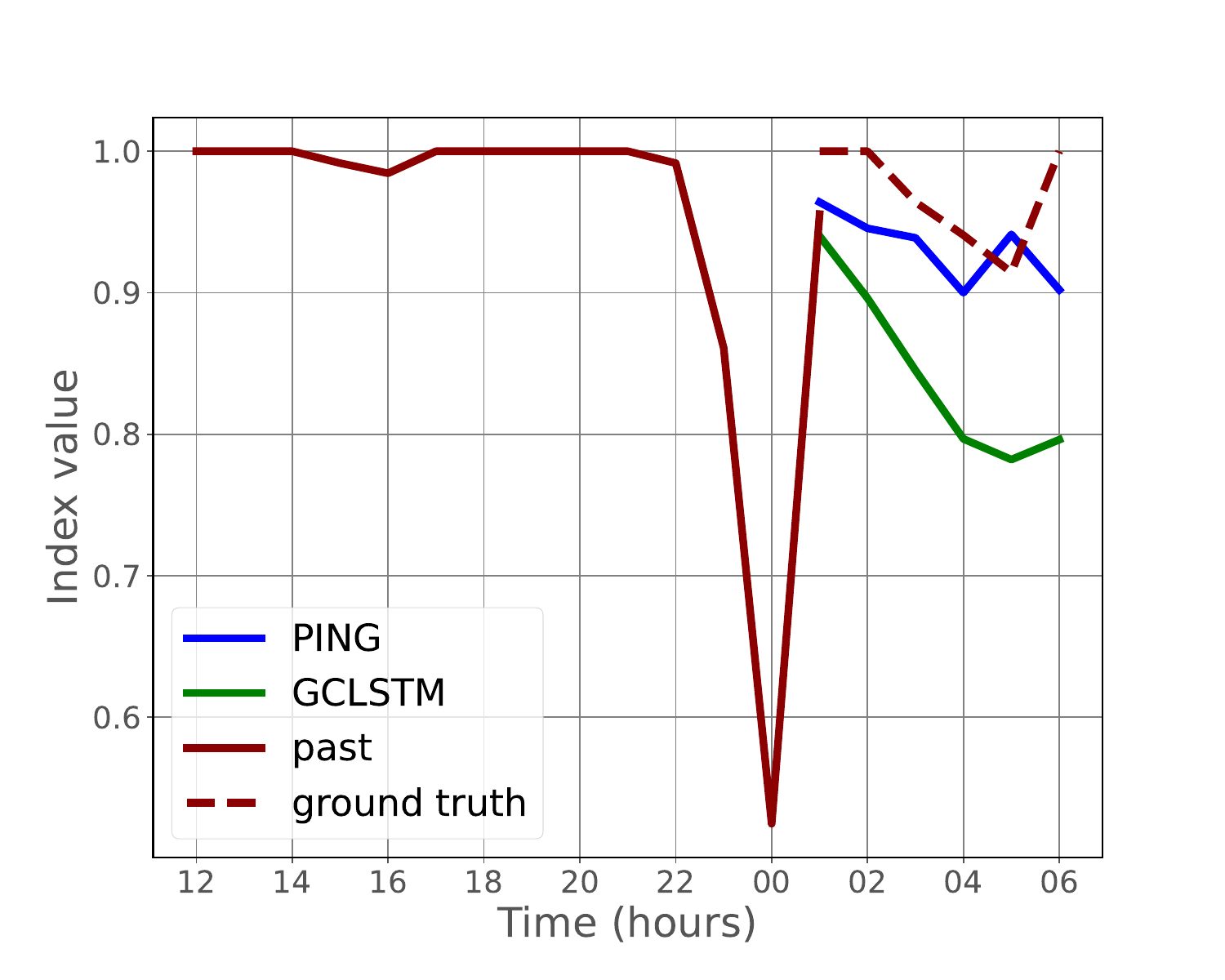}
       \label{fig:ireg21}
  \end{subfigure}
   \caption{ Examples of historical cloud concentration, ground truth and forecasted signals across six forecasting horizon values for a specific node on irregular grid. The results are shown for PING and GCLSTM models.}
  \label{fig:6steps_clouds1}
  \end{figure}

Figure \ref{fig:6steps_clouds1} displays forecast examples made at different times between the 25th and 29th of February on the irregularly sampled grid for the cloud concentration index dataset. These specific instances are selected as they exemplify moments of significant dynamical changes in cloud coverage. Such moments include episodes where the observed value rapidly declines from 1 to 0.55 or exhibits abrupt surges. They are particularly challenging for forecasting models, and as demonstrated in Figure \ref{fig:6steps_clouds1}, PING outperforms GCLSTM for predicting these sudden changes.

\begin{figure}[h!]
  \centering
    \includegraphics[trim=0cm 0cm 0cm 0cm, clip, width=0.7\linewidth]{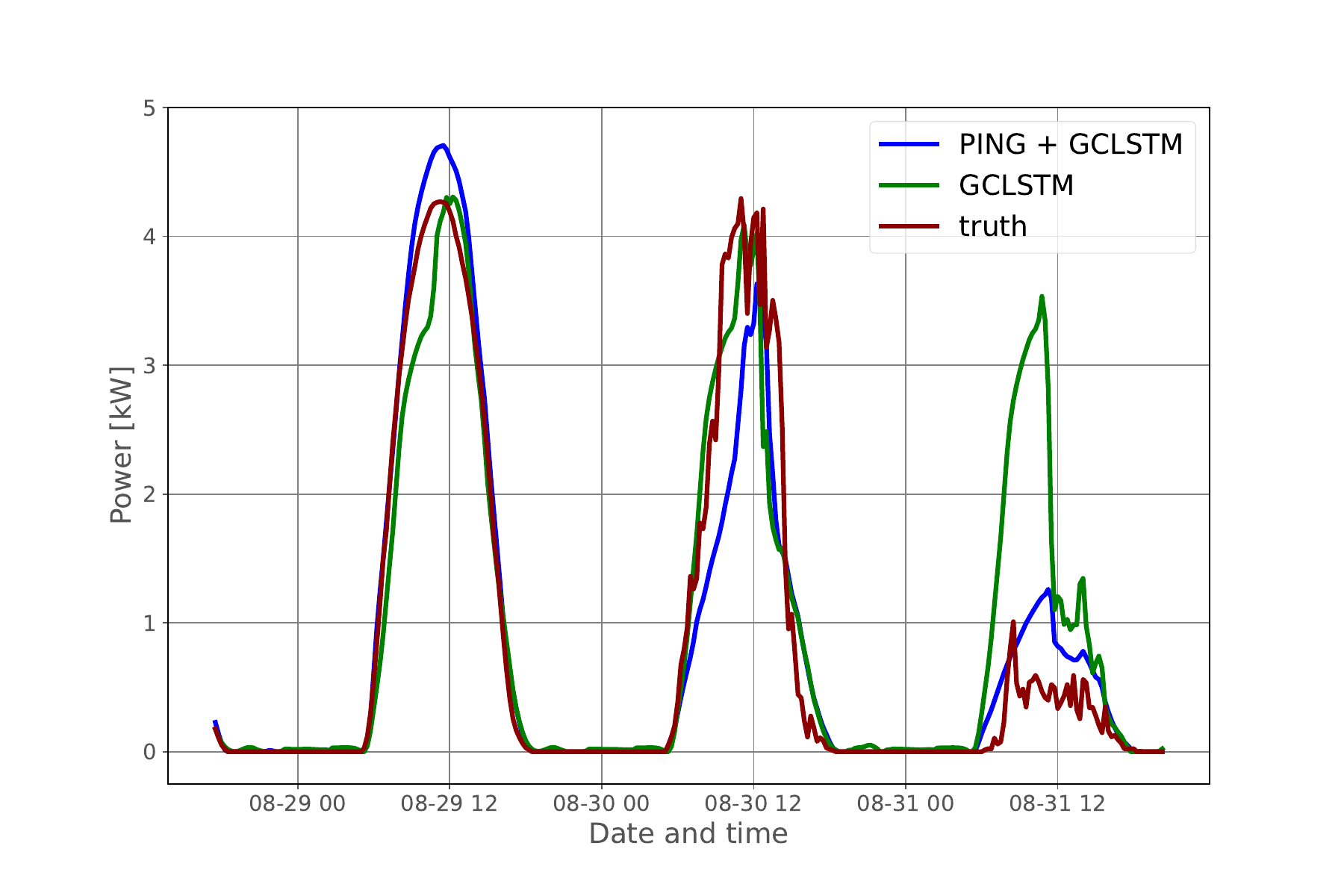}
  \caption{Ground truth and prediction for  sixth hour ahead forecasts during variable days, where each forecast was initialized always six hour before its verification time. }
\label{fig:3days_var}
\end{figure}

 Figure \ref{fig:3days_var} provides examples of forecasts over three consecutive days in August, selected to include sunny, cloudy, and variable conditions. The PING+GCLSTM model performs similarly to the state-of-the-art GCLSTM model during sunny conditions in Figure \ref{fig:3days_var} and outperforms the latter under variable and cloudy conditions (the second and the third day in Figure \ref{fig:3days_var}). The shape of the prediction is closer to the ground truth compared to GCLSTM and  better captures the cloud dynamics. 

 \begin{figure}[h!]
 \centering
    \begin{subfigure}{0.45\linewidth}
    \includegraphics[trim=0cm 0cm 2cm 1.0cm, clip, width=\linewidth]{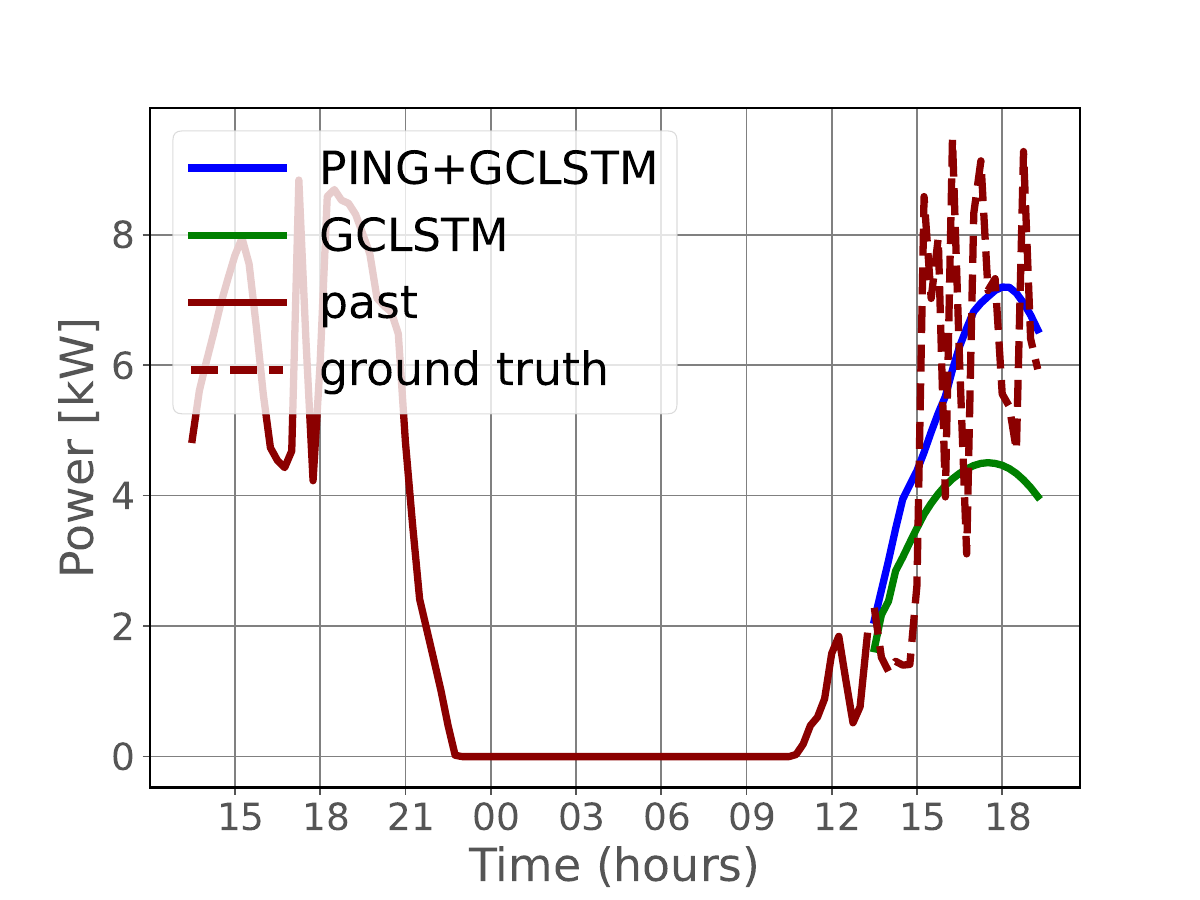}
       \label{fig:ireg211}
  \end{subfigure}
     \begin{subfigure}{0.45\linewidth}
    \includegraphics[trim=0cm 0cm 2cm 1.0cm, clip, width=\linewidth]{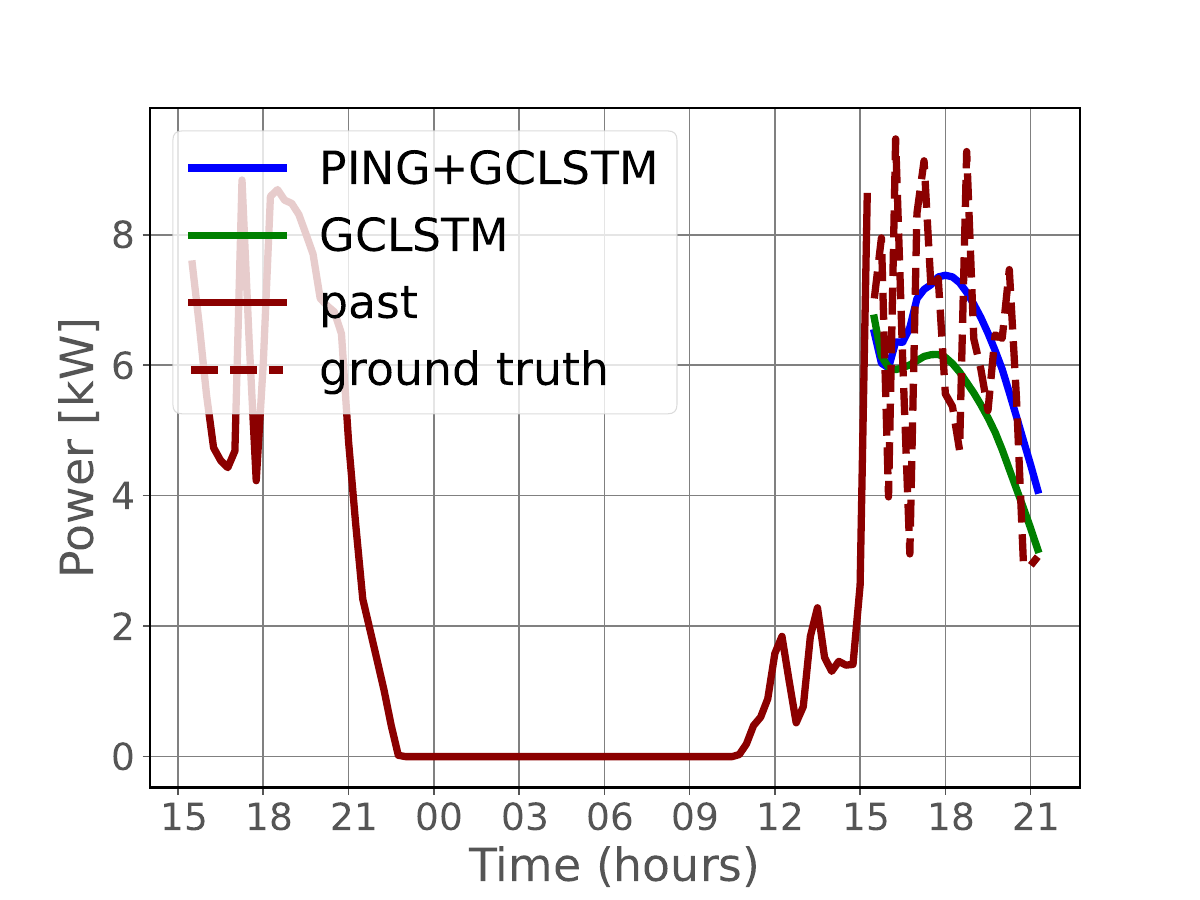}
       \label{fig:ireg311}
  \end{subfigure}
  \caption{Examples of  six hours ahead forecasts initialized at 13:00 and16:00  for the PV power prediction dataset. The forecasted signals and the past signal values are shown for PING+GCLSTM and GCLSTM alone.}
\label{fig:6steps_12}
\end{figure} 

Further analysis compares forecasts made at different times of the day for twenty-four horizon values, during a variable day in Figure \ref{fig:6steps_12}. We have compared the combination of PING and GCLSTM, and GCLSTM. The proposed combination of PING and GCLSTM consistently outperforms GCLSTM by better capturing and predicting the cloud dynamics, particularly early in the morning. GCLSTM relies heavily on clear-sky profiles and local neighborhood information, which can lead to significant errors during cloudy mornings. By incorporating cloud dynamics through PING, the combined model leverages this information from early in the day, resulting in more accurate forecasts. These examples show that PING improved the forecasts when the weather conditions were variable and cloudy. During sunny days, the irradiance is more deterministic, thus all models perform similarly. This highlights the value of physics-informed model when forecasting phenomena that are highly related to cloud dynamics.

\bibliography{biblio_essay.bib}

\begin{thebibliography}{30}
\expandafter\ifx\csname natexlab\endcsname\relax\def\natexlab#1{#1}\fi
\providecommand{\url}[1]{\texttt{#1}}
\providecommand{\href}[2]{#2}
\providecommand{\path}[1]{#1}
\providecommand{\DOIprefix}{doi:}
\providecommand{\ArXivprefix}{arXiv:}
\providecommand{\URLprefix}{URL: }
\providecommand{\Pubmedprefix}{pmid:}
\providecommand{\doi}[1]{\href{http://dx.doi.org/#1}{\path{#1}}}
\providecommand{\Pubmed}[1]{\href{pmid:#1}{\path{#1}}}
\providecommand{\bibinfo}[2]{#2}
\ifx\xfnm\relax \def\xfnm[#1]{\unskip,\space#1}\fi
\bibitem[{Andrychowicz et~al.(2023)Andrychowicz, Espeholt, Li, Merchant,
  Merose, Zyda, Agrawal \& Kalchbrenner}]{andrychowicz2023deep}
\bibinfo{author}{Andrychowicz, M.}, \bibinfo{author}{Espeholt, L.},
  \bibinfo{author}{Li, D.}, \bibinfo{author}{Merchant, S.},
  \bibinfo{author}{Merose, A.}, \bibinfo{author}{Zyda, F.},
  \bibinfo{author}{Agrawal, S.}, \& \bibinfo{author}{Kalchbrenner, N.}
  (\bibinfo{year}{2023}).
\newblock \bibinfo{title}{Deep learning for day forecasts from sparse
  observations}.
\newblock {\it \bibinfo{journal}{arXiv preprint arXiv:2306.06079}\/}, .
\bibitem[{Belbute-Peres et~al.(2020)Belbute-Peres, Economon \&
  Kolter}]{belbute2020combining}
\bibinfo{author}{Belbute-Peres, F. D.~A.}, \bibinfo{author}{Economon, T.}, \&
  \bibinfo{author}{Kolter, Z.} (\bibinfo{year}{2020}).
\newblock \bibinfo{title}{Combining differentiable pde solvers and graph neural
  networks for fluid flow prediction}.
\newblock In {\it \bibinfo{booktitle}{International Conference on Machine
  Learning (ICML)}\/} (pp. \bibinfo{pages}{2402--2411}).
\newblock \bibinfo{organization}{PMLR}.
\bibitem[{Boussif et~al.(2022)Boussif, Bengio, Benabbou \&
  Assouline}]{boussif2022magnet}
\bibinfo{author}{Boussif, O.}, \bibinfo{author}{Bengio, Y.},
  \bibinfo{author}{Benabbou, L.}, \& \bibinfo{author}{Assouline, D.}
  (\bibinfo{year}{2022}).
\newblock \bibinfo{title}{Magnet: Mesh agnostic n pde solver}.
\newblock {\it \bibinfo{journal}{Advances in Neural Information Processing
  Systems (NeurIPS)}\/},  {\it \bibinfo{volume}{35}\/},
  \bibinfo{pages}{31972--31985}.
\bibitem[{Carrillo et~al.(2023)Carrillo, Schubnel, Langou \& Alet}]{clc_rcr}
\bibinfo{author}{Carrillo, R.}, \bibinfo{author}{Schubnel, B.},
  \bibinfo{author}{Langou, R.}, \& \bibinfo{author}{Alet, P.-J.}
  (\bibinfo{year}{2023}).
\newblock \bibinfo{title}{Dynamic graph machine learning for multi-site solar
  forecasting}.
\newblock In {\it \bibinfo{booktitle}{40th European Photovoltaic Solar Energy
  Conference and Exhibition (EUPVSEC), Lisbon, Portugal}\/}.
\bibitem[{Cuomo et~al.(2022)Cuomo, Cola, Giampaolo, Rozza, Raissi \&
  Piccialli}]{Cuomo2022ScientificML}
\bibinfo{author}{Cuomo, S.}, \bibinfo{author}{Cola, V. S.~D.},
  \bibinfo{author}{Giampaolo, F.}, \bibinfo{author}{Rozza, G.},
  \bibinfo{author}{Raissi, M.}, \& \bibinfo{author}{Piccialli, F.}
  (\bibinfo{year}{2022}).
\newblock \bibinfo{title}{Scientific machine learning through
  physics–informed neural networks: Where we are and what’s next}.
\newblock {\it \bibinfo{journal}{Journal of Scientific Computing}\/},  {\it
  \bibinfo{volume}{92}\/}.
\bibitem[{De~B{\'e}zenac et~al.(2019)De~B{\'e}zenac, Pajot \&
  Gallinari}]{de2019deep}
\bibinfo{author}{De~B{\'e}zenac, E.}, \bibinfo{author}{Pajot, A.}, \&
  \bibinfo{author}{Gallinari, P.} (\bibinfo{year}{2019}).
\newblock \bibinfo{title}{Deep learning for physical processes: Incorporating
  prior scientific knowledge}.
\newblock {\it \bibinfo{journal}{Journal of Statistical Mechanics: Theory and
  Experiment}\/},  {\it \bibinfo{volume}{2019}\/}, \bibinfo{pages}{124009}.
\bibitem[{Farneb{\"a}ck(2003)}]{farneback2003two}
\bibinfo{author}{Farneb{\"a}ck, G.} (\bibinfo{year}{2003}).
\newblock \bibinfo{title}{Two-frame motion estimation based on polynomial
  expansion}.
\newblock In {\it \bibinfo{booktitle}{Image Analysis: 13th Scandinavian
  Conference, SCIA 2003 Halmstad, Sweden, June 29--July 2, 2003 Proceedings
  13}\/} (pp. \bibinfo{pages}{363--370}).
\newblock \bibinfo{organization}{Springer}.
\bibitem[{Fortunato et~al.(2022)Fortunato, Pfaff, Wirnsberger, Pritzel \&
  Battaglia}]{fortunato2022multiscale}
\bibinfo{author}{Fortunato, M.}, \bibinfo{author}{Pfaff, T.},
  \bibinfo{author}{Wirnsberger, P.}, \bibinfo{author}{Pritzel, A.}, \&
  \bibinfo{author}{Battaglia, P.} (\bibinfo{year}{2022}).
\newblock \bibinfo{title}{Multiscale meshgraphnets}.
\newblock {\it \bibinfo{journal}{2nd AI4Science Workshop at the 39 th
  International Conference on Machine Learning (ICML)}\/}, .
\bibitem[{Gao et~al.(2022)Gao, Zahr \& Wang}]{GAO2022114502}
\bibinfo{author}{Gao, H.}, \bibinfo{author}{Zahr, M.~J.}, \&
  \bibinfo{author}{Wang, J.-X.} (\bibinfo{year}{2022}).
\newblock \bibinfo{title}{Physics-informed graph neural galerkin networks: A
  unified framework for solving pde-governed forward and inverse problems}.
\newblock {\it \bibinfo{journal}{Computer Methods in Applied Mechanics and
  Engineering}\/},  {\it \bibinfo{volume}{390}\/}, \bibinfo{pages}{114502}.
\bibitem[{He et~al.(2022)He, Li, Zhu \& Tsui}]{9733962}
\bibinfo{author}{He, Y.}, \bibinfo{author}{Li, L.}, \bibinfo{author}{Zhu, X.},
  \& \bibinfo{author}{Tsui, K.~L.} (\bibinfo{year}{2022}).
\newblock \bibinfo{title}{Multi-graph convolutional-recurrent neural network
  (mgc-rnn) for short-term forecasting of transit passenger flow}.
\newblock {\it \bibinfo{journal}{IEEE Transactions on Intelligent
  Transportation Systems}\/},  {\it \bibinfo{volume}{23}\/},
  \bibinfo{pages}{18155--18174}. \DOIprefix\doi{10.1109/TITS.2022.3150600}.
\bibitem[{Holl et~al.(2020)Holl, Koltun \& Thuerey}]{holl2020learning}
\bibinfo{author}{Holl, P.}, \bibinfo{author}{Koltun, V.}, \&
  \bibinfo{author}{Thuerey, N.} (\bibinfo{year}{2020}).
\newblock \bibinfo{title}{Learning to control pdes with differentiable
  physics}.
\newblock {\it \bibinfo{journal}{International Conference on Learning
  Representations (ICLR)}\/}, .
\bibitem[{Lam et~al.(2023)Lam, Sanchez-Gonzalez, Willson, Wirnsberger,
  Fortunato, Pritzel, Ravuri, Ewalds, Alet, Eaton-Rosen
  et~al.}]{lam2022graphcast}
\bibinfo{author}{Lam, R.}, \bibinfo{author}{Sanchez-Gonzalez, A.},
  \bibinfo{author}{Willson, M.}, \bibinfo{author}{Wirnsberger, P.},
  \bibinfo{author}{Fortunato, M.}, \bibinfo{author}{Pritzel, A.},
  \bibinfo{author}{Ravuri, S.}, \bibinfo{author}{Ewalds, T.},
  \bibinfo{author}{Alet, F.}, \bibinfo{author}{Eaton-Rosen, Z.} et~al.
  (\bibinfo{year}{2023}).
\newblock \bibinfo{title}{Graphcast: Learning skillful medium-range global
  weather forecasting}.
\newblock {\it \bibinfo{journal}{Science}\/}, .
\bibitem[{M{\"u}ller \& Remund(2014)}]{satelliteremund}
\bibinfo{author}{M{\"u}ller, S.~C.}, \& \bibinfo{author}{Remund, J.}
  (\bibinfo{year}{2014}).
\newblock \bibinfo{title}{Satellite based shortest term solar energy forecast
  system for entire europe for the next hours}.
\newblock In {\it \bibinfo{booktitle}{30th European Photovoltaic Solar Energy
  Conference and Exhibition (EUPVSEC), Paris, France}\/}.
\bibitem[{Ortega et~al.(2018)Ortega, Frossard, Kova{\v{c}}evi{\'c}, Moura \&
  Vandergheynst}]{8347162}
\bibinfo{author}{Ortega, A.}, \bibinfo{author}{Frossard, P.},
  \bibinfo{author}{Kova{\v{c}}evi{\'c}, J.}, \bibinfo{author}{Moura, J.~M.}, \&
  \bibinfo{author}{Vandergheynst, P.} (\bibinfo{year}{2018}).
\newblock \bibinfo{title}{Graph signal processing: Overview, challenges, and
  applications}.
\newblock {\it \bibinfo{journal}{Proceedings of the IEEE}\/},  {\it
  \bibinfo{volume}{106}\/}, \bibinfo{pages}{808--828}.
\bibitem[{Pan et~al.(2025)Pan, Xu \& Chen}]{PAN2025132572}
\bibinfo{author}{Pan, Z.}, \bibinfo{author}{Xu, L.}, \& \bibinfo{author}{Chen,
  N.} (\bibinfo{year}{2025}).
\newblock \bibinfo{title}{Combining graph neural network and convolutional lstm
  network for multistep soil moisture spatiotemporal prediction}.
\newblock {\it \bibinfo{journal}{Journal of Hydrology}\/},  {\it
  \bibinfo{volume}{651}\/}, \bibinfo{pages}{132572}. \URLprefix
  \url{https://www.sciencedirect.com/science/article/pii/S0022169424019681}.
  \DOIprefix\doi{https://doi.org/10.1016/j.jhydrol.2024.132572}.
\bibitem[{Pfaff et~al.(2020)Pfaff, Fortunato, Sanchez-Gonzalez \&
  Battaglia}]{pfaff2020learning}
\bibinfo{author}{Pfaff, T.}, \bibinfo{author}{Fortunato, M.},
  \bibinfo{author}{Sanchez-Gonzalez, A.}, \& \bibinfo{author}{Battaglia, P.~W.}
  (\bibinfo{year}{2020}).
\newblock \bibinfo{title}{Learning mesh-based simulation with graph networks}.
\newblock {\it \bibinfo{journal}{International Conference on Learning
  Representations (ICLR)}\/}, .
\bibitem[{Raissi et~al.(2019)Raissi, Perdikaris \& Karniadakis}]{raissi}
\bibinfo{author}{Raissi, M.}, \bibinfo{author}{Perdikaris, P.}, \&
  \bibinfo{author}{Karniadakis, G.} (\bibinfo{year}{2019}).
\newblock \bibinfo{title}{Physics-informed neural networks: A deep learning
  framework for solving forward and inverse problems involving nonlinear
  partial differential equations}.
\newblock {\it \bibinfo{journal}{Journal of Computational Physics}\/},  {\it
  \bibinfo{volume}{378}\/}, \bibinfo{pages}{686--707}.
\bibitem[{Sanchez-Gonzalez et~al.(2020)Sanchez-Gonzalez, Godwin, Pfaff, Ying,
  Leskovec \& Battaglia}]{sanchez2020learning}
\bibinfo{author}{Sanchez-Gonzalez, A.}, \bibinfo{author}{Godwin, J.},
  \bibinfo{author}{Pfaff, T.}, \bibinfo{author}{Ying, R.},
  \bibinfo{author}{Leskovec, J.}, \& \bibinfo{author}{Battaglia, P.}
  (\bibinfo{year}{2020}).
\newblock \bibinfo{title}{Learning to simulate complex physics with graph
  networks}.
\newblock In {\it \bibinfo{booktitle}{International Conference on Machine
  Learning (ICML)}\/} (pp. \bibinfo{pages}{8459--8468}).
\newblock \bibinfo{organization}{PMLR}.
\bibitem[{Shi et~al.(2024)Shi, Wang, Qu \& Shao}]{shi2024time}
\bibinfo{author}{Shi, J.}, \bibinfo{author}{Wang, S.}, \bibinfo{author}{Qu,
  P.}, \& \bibinfo{author}{Shao, J.} (\bibinfo{year}{2024}).
\newblock \bibinfo{title}{Time series prediction model using lstm-transformer
  neural network for mine water inflow}.
\newblock {\it \bibinfo{journal}{Scientific reports}\/},  {\it
  \bibinfo{volume}{14}\/}, \bibinfo{pages}{18284}.
\bibitem[{Shi et~al.(2015)Shi, Chen, Wang, Yeung, Wong \&
  Woo}]{shi2015convolutional}
\bibinfo{author}{Shi, X.}, \bibinfo{author}{Chen, Z.}, \bibinfo{author}{Wang,
  H.}, \bibinfo{author}{Yeung, D.-Y.}, \bibinfo{author}{Wong, W.-K.}, \&
  \bibinfo{author}{Woo, W.-c.} (\bibinfo{year}{2015}).
\newblock \bibinfo{title}{Convolutional lstm network: A machine learning
  approach for precipitation nowcasting}.
\newblock {\it \bibinfo{journal}{Advances in neural information processing
  systems}\/},  {\it \bibinfo{volume}{28}\/}.
\bibitem[{Simeunović(2024)}]{infoepfl}
\bibinfo{author}{Simeunović, J.} (\bibinfo{year}{2024}).
\newblock \bibinfo{title}{Network time series forecasting in photovoltaics
  power production}.
\newblock In {\it
  \bibinfo{booktitle}{https://infoscience.epfl.ch/handle/20.500.14299/207077}\/}.
\newblock \bibinfo{publisher}{EPFL}.
\bibitem[{Simeunović et~al.(2022{\natexlab{a}})Simeunović, Schubnel, Alet \&
  Carrillo}]{simeunovic2021spatio}
\bibinfo{author}{Simeunović, J.}, \bibinfo{author}{Schubnel, B.},
  \bibinfo{author}{Alet, P.-J.}, \& \bibinfo{author}{Carrillo, R.~E.}
  (\bibinfo{year}{2022}{\natexlab{a}}).
\newblock \bibinfo{title}{Spatio-temporal graph neural networks for multi-site
  pv power forecasting}.
\newblock {\it \bibinfo{journal}{IEEE Transactions on Sustainable Energy}\/},
  {\it \bibinfo{volume}{13}\/}, \bibinfo{pages}{1210--1220}.
\bibitem[{Simeunović et~al.(2022{\natexlab{b}})Simeunović, Schubnel, Alet,
  Carrillo \& Frossard}]{simeunovic2022interpretable}
\bibinfo{author}{Simeunović, J.}, \bibinfo{author}{Schubnel, B.},
  \bibinfo{author}{Alet, P.-J.}, \bibinfo{author}{Carrillo, R.~E.}, \&
  \bibinfo{author}{Frossard, P.} (\bibinfo{year}{2022}{\natexlab{b}}).
\newblock \bibinfo{title}{Interpretable temporal-spatial graph attention
  network for multi-site pv power forecasting}.
\newblock {\it \bibinfo{journal}{Applied Energy}\/},  {\it
  \bibinfo{volume}{327}\/}, \bibinfo{pages}{120127}.
\bibitem[{Stocker(2011)}]{stocker2011introduction}
\bibinfo{author}{Stocker, T.} (\bibinfo{year}{2011}).
\newblock {\it \bibinfo{title}{Introduction to Climate Modelling}\/}.
\newblock Advances in Geophysical and Environmental Mechanics and Mathematics.
\newblock \bibinfo{publisher}{Springer Berlin Heidelberg}.
\bibitem[{Takeishi \& Kalousis()}]{NEURIPS2021_7ca57a9f}
\bibinfo{author}{Takeishi, N.}, \& \bibinfo{author}{Kalousis, A.} ().
\newblock In {\it \bibinfo{booktitle}{Advances in Neural Information Processing
  Systems (NeurIPS)}\/}.
\bibitem[{Wang \& Yu(2021)}]{wang2021physics}
\bibinfo{author}{Wang, R.}, \& \bibinfo{author}{Yu, R.} (\bibinfo{year}{2021}).
\newblock \bibinfo{title}{Physics-guided deep learning for dynamical systems: A
  survey}.
\newblock {\it \bibinfo{journal}{arXiv preprint arXiv:2107.01272}\/}, .
\bibitem[{Yang et~al.(2025)Yang, Liu, Zhang, Shu \& Zheng}]{YANG2025124744}
\bibinfo{author}{Yang, Y.}, \bibinfo{author}{Liu, Y.}, \bibinfo{author}{Zhang,
  Y.}, \bibinfo{author}{Shu, S.}, \& \bibinfo{author}{Zheng, J.}
  (\bibinfo{year}{2025}).
\newblock \bibinfo{title}{Dest-gnn: A double-explored spatio-temporal graph
  neural network for multi-site intra-hour pv power forecasting}.
\newblock {\it \bibinfo{journal}{Applied Energy}\/},  {\it
  \bibinfo{volume}{378}\/}, \bibinfo{pages}{124744}.
\bibitem[{Yin et~al.(2021)Yin, Le~Guen, Dona, de~B{\'e}zenac, Ayed, Thome \&
  Gallinari}]{yin2021augmenting}
\bibinfo{author}{Yin, Y.}, \bibinfo{author}{Le~Guen, V.},
  \bibinfo{author}{Dona, J.}, \bibinfo{author}{de~B{\'e}zenac, E.},
  \bibinfo{author}{Ayed, I.}, \bibinfo{author}{Thome, N.}, \&
  \bibinfo{author}{Gallinari, P.} (\bibinfo{year}{2021}).
\newblock \bibinfo{title}{Augmenting physical models with deep networks for
  complex dynamics forecasting}.
\newblock {\it \bibinfo{journal}{Journal of Statistical Mechanics: Theory and
  Experiment}\/},  {\it \bibinfo{volume}{2021}\/}, \bibinfo{pages}{124012}.
\bibitem[{Zhang et~al.(2022)Zhang, Sun, Feng, Zhen, Wang, Li, Liu \&
  Wang}]{9773879}
\bibinfo{author}{Zhang, M.}, \bibinfo{author}{Sun, Y.}, \bibinfo{author}{Feng,
  C.}, \bibinfo{author}{Zhen, Z.}, \bibinfo{author}{Wang, F.},
  \bibinfo{author}{Li, G.}, \bibinfo{author}{Liu, D.}, \&
  \bibinfo{author}{Wang, H.} (\bibinfo{year}{2022}).
\newblock \bibinfo{title}{Graph neural network based short-term solar
  irradiance forcasting model considering surrounding meteorological factors}.
\newblock In {\it \bibinfo{booktitle}{2022 IEEE IAS 58th Industrial and
  Commercial Power Systems Technical Conference (I\&CPS)}\/} (pp.
  \bibinfo{pages}{1--9}).
\bibitem[{Zhu et~al.(2023)Zhu, Luo, Huang, Zheng, Su \& Hou}]{zhu2023drcnn}
\bibinfo{author}{Zhu, Y.}, \bibinfo{author}{Luo, S.}, \bibinfo{author}{Huang,
  D.}, \bibinfo{author}{Zheng, W.}, \bibinfo{author}{Su, F.}, \&
  \bibinfo{author}{Hou, B.} (\bibinfo{year}{2023}).
\newblock \bibinfo{title}{Drcnn: decomposing residual convolutional neural
  networks for time series forecasting}.
\newblock {\it \bibinfo{journal}{Scientific Reports}\/},  {\it
  \bibinfo{volume}{13}\/}, \bibinfo{pages}{15901}.

\end{thebibliography}





\end{document}